\documentclass[a4paper,fleqn]{cas-dc}

\usepackage[numbers]{natbib}
\usepackage{color,array}
\usepackage{graphicx}
\usepackage{amsmath}
\usepackage{amssymb}
\usepackage{tabularx}
\usepackage{booktabs}
\usepackage[justification=centering]{caption}
\usepackage{hyperref}
\usepackage{glossaries}

\begin{document}
\let\WriteBookmarks\relax
\def\floatpagepagefraction{1}
\def\textpagefraction{.001}

\shorttitle{Uncertainty-Aware Longitudinal Forecasting of AD Progression}
\shortauthors{A. Hariharan et~al.}

\title[mode=title]{Uncertainty-Aware Longitudinal Forecasting of Alzheimer's Disease Progression Using Deep Learning}

\author[1]{Arya Hariharan}

\ead{arya.hariharan@rvce.edu.in}
\credit{Conceptualization, Methodology, Software, Formal Analysis, Writing -- Original Draft}

\affiliation[1]{organization={Department of Computer Science and Engineering, R.V. College of Engineering},
            city={Bengaluru},
            state={Karnataka},
            country={India}}

\author[2]{Shreyank N. Gowda}
\ead{shreyank.narayanagowda@nottingham.ac.uk}
\credit{Supervision, Writing -- Review \& Editing}
\affiliation[2]{organization={School of Computer Science, University of Nottingham},
            city={Nottingham},
            country={United Kingdom}}

\author[1]{Anala M. R.}
\cormark[1]
\ead{analamr@rvce.edu.in}
\credit{Supervision, Project Administration}

\cortext[1]{Corresponding author}

\begin{abstract}
Longitudinal modelling of Alzheimer’s disease progression is clinically useful only if it can describe not just the most likely next diagnosis, but how a patient may evolve over time and how reliable that forecast is. Most deep learning approaches reduce this problem to single-step classification, treating cognitively normal, mild cognitive impairment, and dementia as flat categories while providing limited insight into how uncertainty accumulates across future visits. We propose a probabilistic framework that combines ordinal diagnosis prediction, multi-horizon trajectory generation, and decomposed uncertainty estimation. A Temporal Fusion Transformer encoder is adapted with a CORAL ordinal output layer, asymmetric loss weighting, and converter oversampling to respect disease-stage ordering and improve sensitivity to MCI-to-dementia transitions. Conditioned on the learned patient-context representation, an autoregressive Mixture Density Network generates five-year probabilistic trajectories for diagnosis state, CDR Sum of Boxes, MMSE orientation, and hippocampal volume. On ADNI, the model outperforms linear, recurrent, and transformer baselines for next-visit diagnosis prediction, with the strongest gains on MCI-versus-dementia discrimination. Generated trajectories achieve near-nominal 90\% credible interval coverage, widening uncertainty across the forecast horizon, and biomarker dynamics consistent with expected Alzheimer’s disease progression. We further separate aleatoric from epistemic uncertainty using analytic mixture variance and a five-member bootstrap ensemble, which provides the strongest encoder diversity and output-level epistemic signal. Epistemic uncertainty is higher for rare progression archetypes, MCI and dementia patients, and under external evaluation on OASIS-3, where it increases alongside prediction error. These results suggest that probabilistic longitudinal forecasting can provide patient-specific disease trajectories together with clinically meaningful signals of when predictions should be treated with caution. Code: \url{https://github.com/Arya-Hari/ldpm-ad}
\end{abstract}

\iffalse
\begin{highlights}
\item Probabilistic longitudinal AD progression framework with ordinal TFT encoder and MDN trajectory generator.
\item Decomposed aleatoric/epistemic uncertainty via law of total variance and bootstrap deep ensemble.
\item Near-nominal 90\% credible interval coverage with widening uncertainty across a 5-year forecast horizon.
\item Epistemic uncertainty correlates with prediction error under OOD transfer to OASIS-3.
\end{highlights}
\fi


\maketitle

\section{Introduction}
\label{sec:introduction}
Alzheimer's disease (AD) is the leading cause of dementia worldwide, affecting an estimated 35 million people and projected to exceed 139 million by 2050 \cite{who2023dementia}. Its clinical course is inherently longitudinal. Progressive
neurodegeneration can begin years or even decades before symptoms become clinically apparent \cite{jack2010hypothetical}, and patients may remain stable, decline slowly, or convert rapidly depending on a complex combination of biological, cognitive, and clinical factors. This temporal nature makes disease progression modelling clinically important. If future trajectories can be characterised reliably from longitudinal
clinical and neuroimaging data, they could support trial enrolment, therapeutic timing, and personalised care planning \cite{reiman2016alzheimer}.

However, the central question in AD progression modelling is not simply what diagnosis a patient will receive at the next visit. A clinically useful model should answer a richer set of questions. How is the patient likely to evolve over the next several years? How wide is the range of plausible futures? Is the model uncertain because the disease course itself is variable, or because the patient is poorly represented in the training data? These questions are especially important for patients with mild cognitive impairment (MCI), whose progression to dementia is heterogeneous and difficult to predict even from
rich longitudinal data \cite{petersen2014mild}.

Existing computational approaches only partially address this problem. Classical
statistical models, including joint models of longitudinal and time-to-event outcomes \cite{rizopoulos2012joint}, event-based models \cite{fonteijn2012event}, and differential equation formulations \cite{oxtoby2017data}, offer useful interpretability but often rely on parametric assumptions that may not capture the heterogeneity of
individual trajectories. Deep learning methods have shown strong discriminative
performance on structured clinical time series \cite{nguyen2020predicting,
ghazi2019training}, but many are still formulated as next-visit classification models, predicting a single future diagnosis label from an observed history. This formulation limits clinical utility in two ways. First, it treats cognitively normal (CN), MCI, and dementia as flat classes, even though AD progression follows a clinically meaningful ordinal structure. A CN-to-MCI error is not equivalent to a CN-to-dementia error, but standard cross-entropy training does not encode this distinction. Second, point prediction provides little insight into the uncertainty of future disease trajectories,
or into whether a model's confidence remains reliable when applied to patients from a different cohort.

Uncertainty quantification has therefore become a central concern in clinical machine learning \cite{kompa2021second, begoli2019need}. In high-stakes prognostic settings, a model that is wrong but confident can be more harmful than one that is wrong but uncertain. For AD, this concern is particularly acute because the most clinically important cases are often the most uncertain: MCI patients may remain stable for years or progress rapidly to dementia, and both outcomes may be plausible from the same observed history. Despite this, most deep learning studies on AD progression report discriminative metrics such as accuracy or AUROC without evaluating how uncertainty propagates across multi-step forecasts, or whether the uncertainty is meaningful under external cohort shift.

A further challenge is that not all uncertainty has the same clinical meaning.
Aleatoric uncertainty reflects the inherent unpredictability of an individual's disease course, while epistemic uncertainty reflects model ignorance that may be reduced with additional or more representative data \cite{hullermeier2021aleatoric, kendall2017uncertainties}. Conflating these two sources can obscure important deployment signals. A patient with high aleatoric uncertainty may require closer monitoring because several disease trajectories remain genuinely plausible. A patient with high epistemic uncertainty may instead indicate that the model has encountered an underrepresented phenotype, missing feature pattern, or cohort shift. This distinction is especially important for models trained on datasets such as the Alzheimer's Disease
Neuroimaging Initiative (ADNI) \cite{weiner2017alzheimer}, which are highly valuable but may not fully represent the diversity of patients encountered in external clinical cohorts.

Motivated by these gaps, we ask four research questions. First, can AD progression be modelled as an ordinal longitudinal process rather than a flat classification problem? Second, can a model generate multi-year probabilistic trajectories rather than only a single next-visit diagnosis? Third, can predictive uncertainty be decomposed into aleatoric and epistemic components that carry different clinical meanings? Fourth, can epistemic uncertainty provide a useful signal when the model is evaluated on an external
cohort unseen during training?

To address these questions, we propose a unified framework for probabilistic
longitudinal AD progression modelling. The framework combines ordinal diagnosis
prediction, autoregressive trajectory generation, and uncertainty decomposition within a single forecasting pipeline. Our primary contributions are as follows:

\begin{enumerate}
\item \textbf{Ordinal Temporal Fusion Transformer.} We adapt the Temporal Fusion
Transformer (TFT) \cite{lim2021temporal}, an architecture designed for multi-horizon forecasting with interpretable attention, to longitudinal AD progression modelling. The TFT is combined with a CORAL ordinal output layer \cite{cao2020rank}, asymmetric loss weighting, and converter oversampling to respect the ordered structure of CN, MCI, and dementia, while prioritising clinically important MCI-to-dementia conversion events.

\item \textbf{Probabilistic trajectory generation.} Conditioned on the TFT
encoder's patient-level context representation, an autoregressive Mixture Density Network (MDN) \cite{bishop1994mixture} generates five-year probabilistic
trajectories over diagnosis state, Clinical Dementia Rating (CDR)-Sum of Boxes,
Mini Mental State Examination (MMSE) orientation, and hippocampal volume. This
allows the model to represent multiple plausible futures rather than a single
deterministic path.

\item \textbf{Decomposed uncertainty estimation.} We decompose predictive
uncertainty into aleatoric and epistemic components using the law of total variance. Aleatoric uncertainty is estimated analytically from the MDN mixture variance, while epistemic uncertainty is estimated using a five-member deep ensemble \cite{lakshminarayanan2017simple}. This enables the model to distinguish between uncertainty arising from intrinsic disease variability and uncertainty arising from limited model knowledge.

\item \textbf{External out-of-distribution validation.} We evaluate the model on
OASIS-3 \cite{lamontagne2019oasis}, a distinct external cohort unseen during
training. We characterise the covariate shift between ADNI and OASIS-3, quantify
the zero-shot transfer gap, and examine whether epistemic uncertainty increases
when predictions become less reliable under distribution shift.

\end{enumerate}

Experiments on ADNI show that the proposed ordinal TFT outperforms linear, recurrent, and Transformer baselines for next-visit diagnosis prediction, with the strongest gains on MCI-versus-dementia discrimination. The trajectory generator produces five-year forecasts with near-nominal 90\% credible interval coverage, widening uncertainty over the forecast horizon, and biomarker dynamics consistent with established AD progression patterns \cite{jack2010hypothetical}. Uncertainty analysis shows that epistemic
uncertainty is higher for rare progression archetypes, MCI and dementia patients, and external OASIS-3 cases where prediction error increases. Together, these findings suggest that probabilistic longitudinal forecasting can provide both patient-specific disease trajectories and clinically meaningful signals of when model predictions should be treated with caution.

The remainder of this paper is organised as follows. Section~\ref{sec:related} reviews related work in disease progression modelling, sequence modelling for clinical data, ordinal learning, and uncertainty quantification. Section~\ref{sec:data} describes the ADNI and OASIS-3 datasets and preprocessing pipeline. Section~\ref{sec:methods} details the model architecture, training procedure, and uncertainty decomposition framework.
Section~\ref{sec:experiments} presents experimental results and ablations.
Section~\ref{sec:discussion} discusses clinical implications, limitations, and directions for future work.

\section{Related Work}
\label{sec:related}

Computational approaches to modelling Alzheimer's disease progression have evolved substantially over the past decade, moving from interpretable statistical models to deep sequence models for longitudinal clinical data. However, four methodological gaps remain central to the present work: most models either impose restrictive assumptions on disease trajectories, focus on single-step deterministic prediction, ignore the ordinal structure of diagnosis transitions, or report uncertainty without separating its clinically distinct sources.

Early work on AD progression established influential biomarker-ordering frameworks such as event-based models \cite{fonteijn2012event} and differential equation formulations \cite{oxtoby2017data}. These approaches remain valuable because they provide interpretable descriptions of disease evolution. However, they typically rely on strong parametric assumptions and may not fully accommodate the heterogeneity observed in large longitudinal cohorts such as ADNI. Deep learning approaches have improved discriminative performance on structured clinical time series \cite{nguyen2020predicting, ghazi2019training}. Recurrent architectures have been used for next-visit diagnosis prediction, while transformer-based models have been explored for imaging-based classification and longitudinal sequence modelling \cite{chen2024adtransformer, tang2025mci}. Despite these advances, much of this work remains centred on predicting the immediately subsequent diagnostic state.

This next-visit classification focus is clinically limiting because it discards long-horizon trajectory information that may be relevant for care planning and trial enrolment. Wang et al.\ \cite{wang2024multimodal} proposed a multimodal deep learning model for long-term AD progression that incorporates interactions between clinical and imaging features, demonstrating improved multi-step prediction over single-modality baselines. Hashemifar et al.\ \cite{hashemifar2022deepad} addressed the single-cohort limitation by training across multiple datasets. He et al.\ \cite{he2025moe} proposed a stage-aware Mixture of Experts framework for neurodegenerative progression modelling using graph neural diffusion, achieving interpretable stage-specific mechanisms. However, these works remain largely deterministic or focus on spatial propagation patterns rather than probabilistic clinical trajectory generation. They do not jointly model multi-year future trajectories with calibrated and decomposed uncertainty estimates, which is the primary gap addressed in this work.

Longitudinal clinical cohorts also pose challenges that are shared with electronic health record data: visits are irregularly spaced, observations may be missing, and patient populations are heterogeneous across sites and protocols. Standard recurrent architectures such as GRUs and LSTMs can process sequential histories, but they often rely on imputation and masking, and may treat temporal gaps uniformly. The Temporal Fusion Transformer (TFT) \cite{lim2021temporal} is well suited to this setting because it jointly models static covariates, observed time-varying features, and known future inputs through variable selection networks and interpretable multi-head attention. TFT-based models have been applied to clinical forecasting tasks such as vital sign prediction in intensive care \cite{phetrittikun2023tft} and multi-modal chest X-ray trajectory prediction \cite{arora2025cxrtft}. ChronoFormer \cite{chronoformer2025} and related time-aware transformer architectures further show the value of continuous-time positional encodings, which modulate attention based on inter-event intervals rather than integer time indices. This is particularly relevant for ADNI, where visit spacing varies from six to twelve months across protocols and phases. However, transformer architectures applied to AD progression modelling have largely been constrained to cross-sectional or two-timepoint inputs \cite{tang2025mci}, and have not been widely adapted to generate probabilistic multi-step forecasts. Our work addresses this gap by adapting the TFT architecture to the ADNI longitudinal setting as a probabilistic encoder whose patient-level context representations condition a generative trajectory model.

A second limitation is that disease severity in AD is inherently ordinal. The CN-MCI-Dementia spectrum admits a natural clinical ordering, but standard multi-class cross-entropy treats all diagnostic classes as unrelated and all misclassification errors as equally distant. Ordinal regression frameworks for deep neural networks explicitly encode this ordering through cumulative link models and ranked output layers. CORAL \cite{cao2020rank} achieves rank consistency by sharing weights across binary threshold classifiers, while CORN \cite{shi2023corn} relaxes the weight-sharing constraint while preserving consistency through a conditional chain-rule formulation. Ordinal learning has also been adopted in medical staging tasks beyond AD. Uncertainty-aware ordinal networks have been applied to diabetic retinopathy grading \cite{ordinaldr2026}, and Kamal and Farooq \cite{kamal2024ordinalreslogit} integrated CORAL into a residual logit architecture. Despite these advances, ordinal learning in disease progression modelling has mostly been applied to single-step classification. In this work, we extend ordinal constraints to both the classification head and the autoregressive transition function of a generative model, ensuring that generated trajectories respect the biological ordering of disease states by construction.

Generating plausible future trajectories rather than point predictions requires a model capable of representing multi-modal outcome distributions. This is especially important for MCI patients, whose long-term prognosis may include stability or conversion to dementia at varying rates. Mixture Density Networks (MDNs) \cite{bishop1994mixture} address this by parameterising a Gaussian mixture model at each output step, allowing multi-modal conditional distributions to be represented without discrete mode assignment. MDNs have been applied in epidemiological modelling to emulate stochastic within-host models and complex individual-based simulations \cite{carruthers2023mdn}, demonstrating their ability to capture distributional diversity in sequential biological data. For disease progression, probabilistic mixture extensions of mixed-effects models have been proposed for identifying clinically meaningful subtypes \cite{mixture2025subtypes}, while Hidden Markov model variants have modelled heterogeneous progression as transitions between latent states \cite{iohmm2022}. These approaches offer interpretability, but are often limited to low-dimensional state spaces and do not scale naturally to generative tasks conditioned on rich neuroimaging histories. Our MDN autoregressive generator addresses this by conditioning generation on the TFT context vector, enabling coherent multi-biomarker trajectory sampling across time through a GRU hidden state.

Uncertainty quantification has emerged as a central concern for deploying machine learning in clinical settings \cite{kompa2021second, begoli2019need}. A clinician interacting with a prognostic AI system needs to know not only what the model predicts, but how reliable the prediction is and why the model is uncertain. Recent surveys \cite{uqhealthcare2025, oodmedical2024} identify deep ensembles \cite{lakshminarayanan2017simple}, Monte Carlo dropout, and Bayesian neural networks as widely used methods for estimating predictive uncertainty in medical AI. Deep ensembles in particular have been shown to produce well-calibrated uncertainty estimates and strong performance under distribution shift \cite{ensembles2025frequentist}. A critical distinction is between aleatoric uncertainty, which captures irreducible data noise, and epistemic uncertainty, which captures reducible model uncertainty \cite{hullermeier2021aleatoric, kendall2017uncertainties}. Koch et al.\ \cite{koch2024distribution} demonstrate that distribution shifts in deployed medical AI systems can be detected through changes in model uncertainty, motivating epistemic uncertainty as a deployment-time signal of out-of-distribution inputs. Weng et al.\ \cite{weng2025ood} similarly frame OOD detection as a risk-control strategy for medical classification models, showing that uncertainty-based detectors can flag shifts arising from biological variability, cohort differences, and missing data.

Despite this growing body of work, decomposed uncertainty estimation for longitudinal disease progression remains underexplored. Existing approaches either report total predictive uncertainty without decomposition, or apply decomposition to single-step classification tasks. Our work fills this gap through the law of total variance, using the MDN mixture variance as an analytic aleatoric estimator and a five-member deep ensemble as the epistemic estimator. This produces per-step, per-biomarker uncertainty decompositions that we validate both within ADNI and under cross-cohort transfer to OASIS-3 \cite{lamontagne2019oasis}.

\section{Data}
\label{sec:data}

\subsection{Primary Dataset: Alzheimer's Disease Neuroimaging Initiative}

The primary data used in this study were obtained from the Alzheimer's Disease Neuroimaging Initiative (ADNI) database (\url{adni.loni.usc.edu}). ADNI is a longitudinal data collection initiative launched in 2003 by the National Institute on Aging (NIA), with the goal of developing validated biomarkers for Alzheimer's disease clinical trials \cite{weiner2017alzheimer}. The initiative has progressed through four phases, ADNI1, ADNIGO, ADNI2, and ADNI3, enrolling participants across sites in the United States and Canada \cite{aisen2024clinicalcore}. In this work, we used the ADNIMERGE2 composite dataset, which combines participant records across ADNI phases into a single longitudinal structure.

\subsubsection{Participants and diagnosis criteria}

Participants were represented using three ordinal diagnostic categories: cognitively normal (CN, $y=0$), mild cognitive impairment (MCI, $y=1$), and dementia ($y=2$). Within ADNI, CN participants show no subjective or objective memory impairment, MCI participants exhibit objective cognitive decline without functional impairment, and the dementia category includes clinically confirmed Alzheimer's-type dementia. This three-level ordinal encoding is consistent with the biomarker-staging model of \cite{jack2010hypothetical} and has been used in prior longitudinal modelling work \cite{nguyen2020predicting}.

\subsubsection{Feature taxonomy}

We defined a feature taxonomy with four groups, selected to capture the main clinical, cognitive, and neuroimaging dimensions of Alzheimer's disease progression while maintaining availability across ADNI phases.

\begin{itemize}
\item \textbf{Static features} (4): biological sex, years of education, APOE4 allele dose (0, 1, or 2 risk alleles), and APOE2 carrier status. These features are constant across visits for each participant.

\item \textbf{MRI features} (8): hippocampal total volume, total lateral ventricular volume, entorhinal total volume, amygdala total volume, all normalised by intracranial volume, and cortical thickness of the left and right entorhinal cortex, left fusiform gyrus, and left inferior temporal gyrus.

\item \textbf{Cognitive features} (6): four MMSE subscores, recall, orientation to time, orientation to place, and attention/calculation, together with ADAS-Cog 11 and ADAS-Cog 13 total scores.

\item \textbf{CDR features} (8): CDR Sum of Boxes (CDR-SB), CDR global score, and the six CDR domain subscores, memory, orientation, judgment and problem-solving, community affairs, home and hobbies, and personal care.

\end{itemize}

The diagnosis label was also included as a time-varying input feature, allowing the model to condition on the observed disease state when predicting future visits. The complete per-visit feature vector therefore had dimensionality $4 + 1 + 8 + 6 + 8 + 1 = 28$, comprising static covariates, time-varying biomarkers, and the diagnosis label.

\subsubsection{Preprocessing}

ADNIMERGE2 is provided through separate clinical tables corresponding to different assessment categories. These tables were merged across ADNI phases using the participant identifier (RID), with a 30-day tolerance window applied through nearest-neighbour matching on visit dates. When multiple assessments occurred within the same tolerance window, typically because of protocol overlap across ADNI phases, the record from the most recent phase was retained.

Exclusion and preprocessing criteria were then applied sequentially. Participants with fewer than three recorded visits were excluded, since very short sequences provide limited longitudinal context for prediction. This criterion is consistent with prior longitudinal deep learning work \cite{ghazi2019training}. Within each participant trajectory, forward and backward filling were used to handle sporadic missing values caused by missed assessments or protocol changes.

\begin{figure*}[h]
\centering
\includegraphics[width=1.0\textwidth]{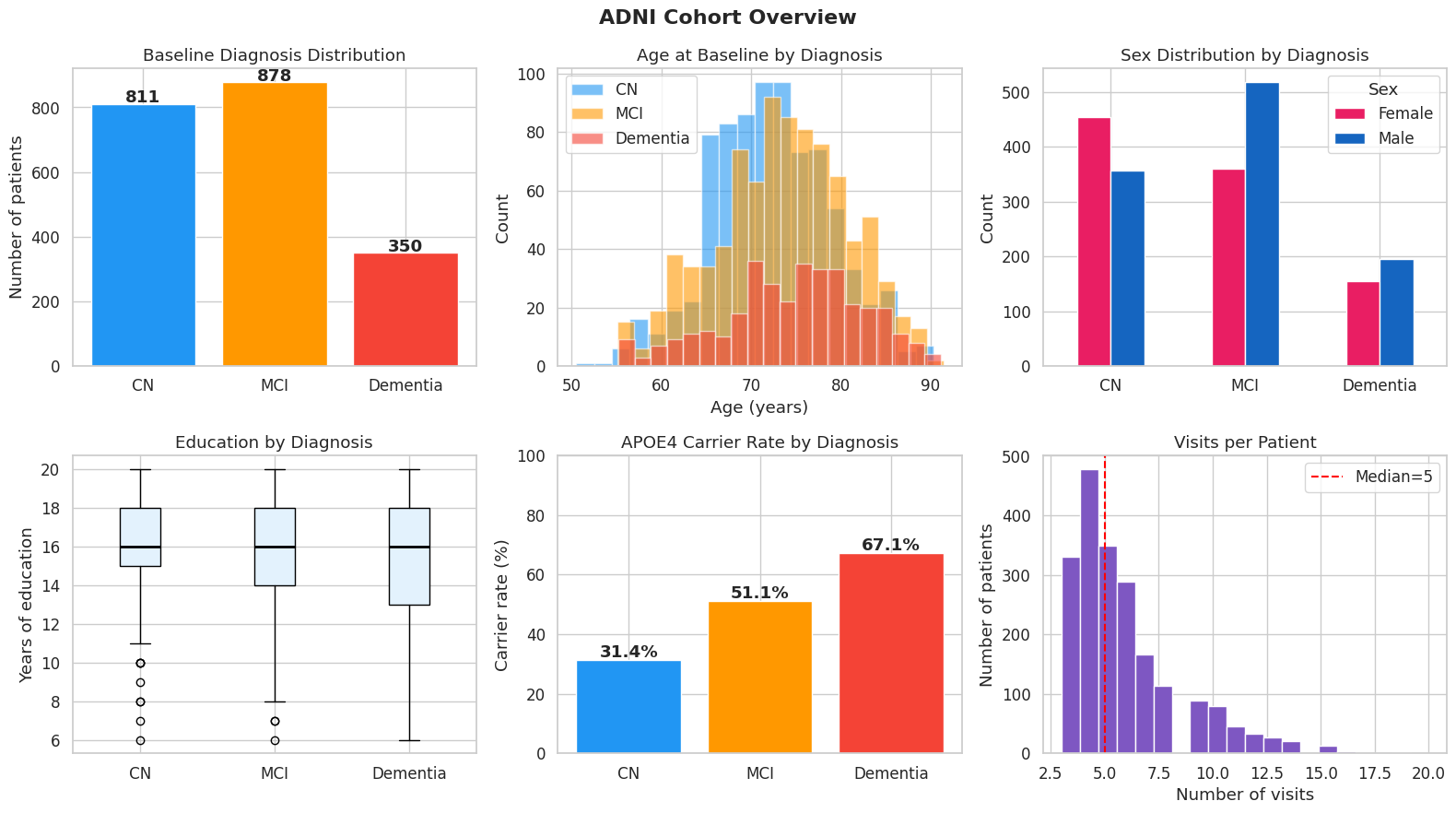}
\caption{Final dataset cohort overview.}
\label{fig:cohort_overview}
\end{figure*}

The final dataset comprised 2039 participants. Figure~\ref{fig:cohort_overview} provides an overview of the resulting cohort.

\subsubsection{Dataset construction}

Participants were split into training (70\%), validation (15\%), and test (15\%) sets using a stratified group split. The participant identifier was used as the grouping variable to prevent data leakage across sequences from the same individual. Stratification was performed using baseline diagnosis, preserving the CN, MCI, and dementia distribution across splits. Temporal sequences were constructed independently within each split. For a participant with $T$ visits, $T-1$ input-target pairs were generated. For each pair, the input consisted of all visits up to time $t$, left-padded to a maximum sequence length of 20, and the target was the diagnosis and biomarker state at visit $t+1$.

\subsection{External Validation Dataset: OASIS-3}

For out-of-distribution (OOD) evaluation, we used OASIS 3 (Open Access Series of Imaging Studies), a retrospective compilation of neuroimaging, clinical, and cognitive data collected over 30 years at the Washington University Knight Alzheimer Disease Research Center \cite{lamontagne2019oasis}. OASIS-3 provides a distinct external cohort, and a random subset of 300 patients was selected for OOD evaluation.

Most features mapped directly between ADNI and OASIS 3, including CDR subscores, FreeSurfer volumetric and surface features, and diagnosis labels. Two feature groups required approximation. First, OASIS-3 records only the MMSE total score rather than the four MMSE subfeatures used in ADNI. These four MMSE subfeatures were therefore imputed using their corresponding training-set means from ADNI. This conservative strategy avoids injecting cohort-specific information into these dimensions. Second, ADAS-Cog is not administered in OASIS-3. Both ADAS-Cog 11 and ADAS-Cog 13 were therefore imputed using ADNI training-set means. No fine-tuning on OASIS-3 was performed at any stage.

\subsection{Ethical Considerations}

All ADNI data were collected under protocols approved by the institutional review boards of each participating site \cite{weiner2017alzheimer}. OASIS-3 data were collected in accordance with Washington University institutional review board protocols. Both datasets are governed by data use agreements that permit academic research use. No patient re-identification was performed in this work.

\section{Methods}
\label{sec:methods}

Let $\mathcal{H}_t = \{(\mathbf{x}_1, \ldots, \mathbf{x}_t), \mathbf{s}\}$ denote the
observed history for a patient at visit $t$, where $\mathbf{x}_\tau \in \mathbb{R}^{24}$
is the time-varying feature vector at visit $\tau$ (MRI biomarkers, cognitive scores,
CDR subscores, and current diagnosis) and $\mathbf{s} \in \mathbb{R}^4$ is the static
covariate vector (sex, education, APOE4 dose, APOE2 carrier status). For this data, we describe two prediction tasks.

\begin{enumerate}
    \item \textbf{Next-visit ordinal classification} Predict the diagnosis label $y_{t+1}
    \in \{0, 1, 2\}$ at the next visit, where the labels encode the ordinal sequence
    CN $<$ MCI $<$ Dementia. The ordinal structure imposes the constraint that
    misclassifying CN as Dementia should be penalised more heavily than misclassifying
    CN as MCI.

    \item \textbf{Probabilistic trajectory generation} Given $\mathcal{H}_t$, generate
    a set of $S$ plausible future trajectories
    $\{\boldsymbol{\tau}^{(s)}\}_{s=1}^{S}$, where each trajectory
    $\boldsymbol{\tau}^{(s)} = \{(y^{(s)}_h, \mathbf{b}^{(s)}_h)\}_{h=1}^{H}$
    specifies a diagnosis state and biomarker vector
    $\mathbf{b}_h = (\text{CDR-SB}_h, \text{MMSE-orient}_h, \text{Hippocampus}_h)$
    at each of $H = 10$ future visits spaced 6 months apart, covering a 5-year horizon.
\end{enumerate}

\begin{figure*}[h]
    \centering
    \includegraphics[width=1.0\textwidth]{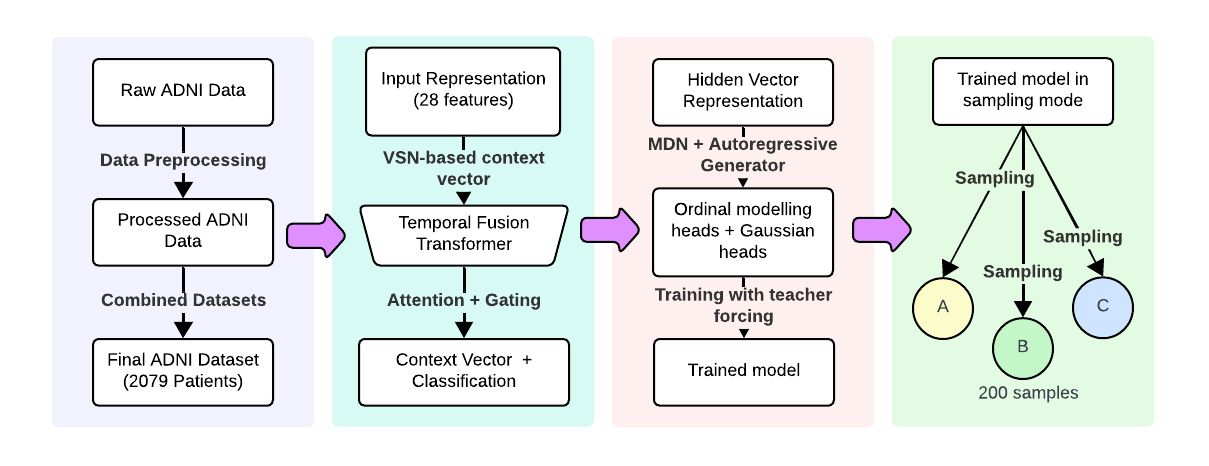}
    \caption{Architecture overview}
    \label{fig:architecture}
\end{figure*}

An overview of the full framework is shown in Figure~\ref{fig:architecture}.

\subsection{Temporal Fusion Transformer Encoder}
\label{sec:tft}

We adapt the Temporal Fusion Transformer \cite{lim2021temporal} as the primary encoder,
mapping the observed history $\mathcal{H}_t$ to a fixed-dimensional context vector
$\mathbf{h} \in \mathbb{R}^{64}$ that encodes patient-level progression state. The
TFT is particularly suited to this setting because it provides native handling of
heterogeneous static and time-varying inputs through separate processing streams, and
its interpretable multi-head attention allows inspection of which past visits drive
predictions.

The encoder consists of five components applied sequentially. Separate VSNs are applied to the static features $\mathbf{s}$ and the time-varying features $\mathbf{x}_\tau$. Each VSN projects individual scalar features into a shared
embedding space, concatenates these
embeddings, and passes them through a Gated Residual Network (GRN) to produce
normalised importance weights $\boldsymbol{\alpha} \in \Delta^{F-1}$ (a probability
simplex over features). The final output is a weighted sum of per-feature embeddings
after an additional per-feature GRN:
\begin{equation}
\begin{aligned}
\text{VSN}(\mathbf{x})
&= \sum_{f=1}^{F} \alpha_f \cdot \text{GRN}_f(\mathbf{e}_f), \\
\alpha_f
&= \text{softmax}\!\left(
\text{GRN}_{\text{flat}}
([\mathbf{e}_1,\ldots,\mathbf{e}_F])
\right)_f
\end{aligned}
\end{equation}
where $\mathbf{e}_f = \mathbf{W}_f x_f$ is the scalar embedding of the $f$-th feature.
The temporal VSN is additionally conditioned on th static context vector $\mathbf{c}_s$. All internal transformations use GRNs, which combine a gated linear unit (GLU) with a
skip connection and layer normalisation:
\begin{equation}
\begin{aligned}
\text{GRN}(\mathbf{x},\mathbf{c})
&= \text{LayerNorm}\Bigl(
\mathbf{x}' + \text{GLU}\bigl(
\mathbf{W}_2 \cdot \text{ELU}(z)
\bigr)
\Bigr),\\
z
&= \mathbf{W}_1\mathbf{x}
+ \mathbf{W}_c\mathbf{c}
+ \mathbf{b}_1
\end{aligned}
\end{equation}
where $\mathbf{x}' = \mathbf{W}_\text{proj}\mathbf{x}$ if $\dim(\mathbf{x}) \neq
\dim(\text{output})$, else $\mathbf{x}' = \mathbf{x}$, and $\mathbf{c}$ is an optional
context vector. The GLU gate $\text{GLU}(\mathbf{z}) = \mathbf{z}_1 \odot
\sigma(\mathbf{z}_2)$ controls information flow, suppressing irrelevant activations. The static embedding $\mathbf{e}_s = \text{VSN}_s(\mathbf{s})$ is projected through
four separate GRNs to produce four context vectors: $\mathbf{c}_s$ (temporal VSN
enrichment), $\mathbf{c}_e$ (static enrichment), $\mathbf{c}_h$ (LSTM hidden state
initialisation), and $\mathbf{c}_c$ (LSTM cell state initialisation). The temporally embedded features are processed by a single-layer LSTM, initialised from $(\mathbf{c}_h, \mathbf{c}_c)$. The LSTM output is
combined with the pre-LSTM embeddings via a GLU gate and layer normalisation, yielding
a sequence of enriched representations $\{\mathbf{e}^\text{enc}_\tau\}_{\tau=1}^{t}$. \\

A modification done relative to the standard multi-head attention \cite{vaswani2017attention}
is the use of a shared value projection across all attention heads:
\begin{equation}
\begin{aligned}
\text{IMHA}(\mathbf{E})
&= \mathbf{W}_O
\left(
\bar{\mathbf{A}}
\cdot
\mathbf{W}_V \mathbf{E}
\right),\\
\bar{\mathbf{A}}
&=
\frac{1}{H_{\text{att}}}
\sum_{h=1}^{H_{\text{att}}}
\text{softmax}\!\left(
\frac{
\mathbf{Q}_h\mathbf{K}_h^\top
}{
\sqrt{d/H_{\text{att}}}
}
\right)
\end{aligned}
\end{equation}
where $H_{\text{att}} = 4$ attention heads, and the averaged attention matrix
$\bar{\mathbf{A}} \in \mathbb{R}^{t \times t}$ can be inspected to identify which
past visits the model attends to when generating a prediction. The attention output is
combined with the LSTM encoding via a second GLU gate and fed through a feed-forward
GRN. The representation at the final time step $t$ is passed through a static
enrichment GRN conditioned on $\mathbf{c}_e$, producing the context vector $\mathbf{h}$. \\

For next-visit classification, we attach a rank-consistent ordinal output layer following the CORAL framework \cite{cao2020rank}. A shared linear projection maps
$\mathbf{h}$ to a scalar:
\begin{equation}
g(\mathbf{h}) = \mathbf{w}_2^\top \text{ReLU}(\mathbf{W}_1 \mathbf{h})
\end{equation}
and $K-1 = 2$ learnable bias parameters $\{b_k\}_{k=0}^{1}$ (initialised at
$[0.5, -0.5]$) define cumulative threshold logits $\ell_k = g(\mathbf{h}) + b_k$.
Cumulative probabilities are $\hat{P}(y > k) = \sigma(\ell_k)$, and class probabilities
are recovered as:
\begin{equation}
\hat{P}(y = k) = \hat{P}(y > k-1) - \hat{P}(y > k), \quad k \in \{0, 1, 2\}
\end{equation}
with $\hat{P}(y > -1) \triangleq 1$ and $\hat{P}(y > K-1) \triangleq 0$. By
construction, $\hat{P}(y = k) \geq 0$ for all $k$ and $\sum_k \hat{P}(y = k) = 1$,
guaranteeing a valid probability distribution that respects the ordinal ordering. The encoder is also trained with a CORAL loss, where binary cross-entropy is applied independently
over each of the $K-1$ cumulative thresholds:
\begin{equation}
\begin{aligned}
\mathcal{L}_{\text{CORAL}}(\mathbf{h},y)
&=
-\frac{1}{K-1}
\sum_{k=0}^{K-2}
\Bigl[
\mathbb{1}[y>k]\log\sigma(\ell_k)
\\
&\qquad\qquad
+
\mathbb{1}[y\le k]
\log\bigl(1-\sigma(\ell_k)\bigr)
\Bigr]
\end{aligned}
\end{equation}
To further improve sensitivity to conversion events, a converter
oversampling strategy is applied during training where transition pairs (two consequent visits where a diagnosis change happens) are randomly oversampled comapred to stable pairs (two consequent visits where diagnosis does not change).

\subsection{Probabilistic Trajectory Generator}
\label{sec:generator}
Conditioned on the frozen TFT encoder context vector $\mathbf{h}$, an autoregressive
trajectory generator produces probabilistic multi-step forecasts over biomarkers and
diagnosis state. The generator is frozen after training and the encoder weights are not
updated during generator training.

At each future step $h \in \{1, \ldots, H\}$, a GRU cell maintains a hidden state
$\mathbf{z}_h \in \mathbb{R}^{64}$ that tracks the evolving trajectory:
\begin{equation}
\mathbf{z}_h = \text{GRU}\bigl(\text{MLP}([\mathbf{h};\, \mathbf{b}_{h-1};\, \mathbf{d}_{h-1};\, \Delta_h]),\, \mathbf{z}_{h-1}\bigr)
\end{equation}
where $\mathbf{b}_{h-1} \in \mathbb{R}^3$ is the previous biomarker state,
$\mathbf{d}_{h-1} \in \{0,1\}^3$ is the one-hot encoding of the previous diagnosis,
$\Delta_h = 6$ months is the fixed inter-visit interval, and $\mathbf{z}_0 =
\mathbf{W}_\text{init}\mathbf{h}$ initialises the GRU from the encoder context. The GRU hidden state $\mathbf{z}_h$ is passed to a Mixture Density Network head
\cite{bishop1994mixture} that parameterises a $K=3$ component Gaussian mixture over the
three biomarker targets:
\begin{align}
\boldsymbol{\pi}_h &= \text{softmax}(\mathbf{W}_\pi \mathbf{z}_h), \quad \boldsymbol{\pi}_h \in \Delta^{K-1} \\
\boldsymbol{\mu}_{h,k} &= \mathbf{W}_\mu^{(k)} \mathbf{z}_h, \quad \boldsymbol{\mu}_{h,k} \in \mathbb{R}^3 \\
\boldsymbol{\sigma}_{h,k} &= \exp(\mathbf{W}_\sigma^{(k)} \mathbf{z}_h), \quad \boldsymbol{\sigma}_{h,k} \in \mathbb{R}_{>0}^3
\end{align}
A biomarker sample is drawn by first selecting a component $k \sim \text{Categorical}
(\boldsymbol{\pi}_h)$. Then sampling is done where $\hat{\mathbf{b}}_h \sim
\mathcal{N}(\boldsymbol{\mu}_{h,k}, \text{diag}(\boldsymbol{\sigma}_{h,k}^2))$.
Sigma values are clamped to $[10^{-4}, 10]$ for numerical stability. \\

Diagnosis at step $h$ is drawn from a learned ordinal transition distribution
$P(y_h \mid y_{h-1}, \mathbf{z}_h)$. A linear layer maps $\mathbf{z}_h$ to
three logits, which are then masked to enforce clinically motivated structural
constraints before softmax:
\begin{equation}
P(y_h = j \mid y_{h-1} = i, \mathbf{z}_h) \propto \exp(\ell_{ij}) \cdot \mathbb{1}[\mathcal{M}_{ij} = 0]
\end{equation}
where $\mathcal{M} \in \{0,1\}^{3 \times 3}$ is a hard impossibility mask with
$\mathcal{M}_{02} = \mathcal{M}_{10} = \mathcal{M}_{20} = \mathcal{M}_{21} = 1$.
This prohibits CN$\to$Dementia transitions in a single step,
and all backward transitions.
The mask is applied by setting masked logits to $-\infty$ before softmax, yielding
zero probability for impossible transitions by construction rather than by regularisation. \\

The generator is trained with teacher forcing where at each step, the true biomarker and
diagnosis values from the training sequence are used as the input $(\mathbf{b}_{h-1},
y_{h-1})$ rather than the model's own samples, stabilising early training. The
composite loss is:
\begin{equation}
\begin{aligned}
\mathcal{L}_{\text{gen}}
&=
\frac{1}{H_{\text{valid}}}
\sum_{h=1}^{H_{\text{valid}}}
\Bigl[
-\log p_{\text{MDN}}
\bigl(
\mathbf{b}_h
\mid
\boldsymbol{\pi}_h,
\boldsymbol{\mu}_h,
\boldsymbol{\sigma}_h
\bigr)
\\
&\qquad\qquad
+
\mathcal{L}_{\text{CE}}
\bigl(
y_h,\,
\hat{P}(
\cdot \mid y_{h-1},
\mathbf{z}_h
)
\bigr)
\Bigr]
\end{aligned}
\end{equation}
where $H_\text{valid}$ is the number of valid (non-padded) future steps, and the MDN log-likelihood is $\log p_\text{MDN}(\mathbf{b})
=\log\allowbreak
\sum_{k=1}^{K}\pi_k\allowbreak
\mathcal{N}(\mathbf{b};\allowbreak
\boldsymbol{\mu}_k,\allowbreak
\boldsymbol{\sigma}_k^2\mathbf{I})$

At inference, $S = 200$ trajectories are sampled per patient by running the
autoregressive model stochastically: at each step, a component is sampled from
$\boldsymbol{\pi}_h$, a biomarker vector is sampled from the corresponding Gaussian,
and a diagnosis is sampled from the transition distribution. The sampled values are fed
back as inputs to the next step, without teacher forcing. To identify clinically interpretable progression archetypes, a Variational Autoencoder (VAE) is trained
on the generated samples. Each trajectory is represented as a concatenation of biomarker
values and diagnosis probabilities across all $H$ steps. The VAE encoder maps this to a lower latent space with reparameterised sampling
\cite{kingma2014vae} and a $\beta$-VAE objective \cite{higgins2017betavae} with
$\beta = 0.5$:
\begin{equation}
\mathcal{L}_\text{VAE} = \mathbb{E}_{q_\phi(\mathbf{z}|\mathbf{x})}[\log p_\theta(\mathbf{x}|\mathbf{z})] - \beta \cdot D_\text{KL}(q_\phi(\mathbf{z}|\mathbf{x}) \| \mathcal{N}(\mathbf{0}, \mathbf{I}))
\end{equation}
$K$-means clustering ($K=4$) is applied to the mean latent codes from the training set,
and archetypes are labelled by ascending mean CDR-SB at the final forecast step as:
Stable, Slow Decline, Moderate Decline, and Rapid Progression.

\subsection{Uncertainty Decomposition}
\label{sec:uncertainty}

We decompose total predictive uncertainty into aleatoric and epistemic components using
the law of total variance. For a predicted biomarker value $\hat{b}$ at a given future
step, the decomposition is:
\begin{equation}
\underbrace{\text{Var}[\hat{b}]}_{\text{total}} =
\underbrace{\mathbb{E}_\theta[\text{Var}[\hat{b} \mid \theta]]}_{\text{aleatoric}} +
\underbrace{\text{Var}_\theta[\mathbb{E}[\hat{b} \mid \theta]]}_{\text{epistemic}}
\label{eq:lotv}
\end{equation}
where $\theta$ indexes model parameters. Intuitively, aleatoric uncertainty captures
the expected variance of individual model outputs (irreducible data noise), while
epistemic uncertainty captures the variance of model expectations across the parameter
ensemble (reducible model parameter uncertainty). \\

The aleatoric term in Equation~\ref{eq:lotv} is estimated analytically from the MDN
parameters. For a $K$-component mixture, the total variance of the mixture distribution
is:
\begin{equation}
\text{Var}_\text{alea}[\hat{b}_h] = \underbrace{\sum_{k=1}^{K} \pi_k \sigma_{h,k}^2}_{\mathbb{E}[\text{Var}]} + \underbrace{\sum_{k=1}^{K} \pi_k \mu_{h,k}^2 - \left(\sum_{k=1}^{K} \pi_k \mu_{h,k}\right)^2}_{\text{Var}[\mathbb{E}]}
\end{equation}
This is computed in a single forward pass without sampling, making it an efficient and
analytically exact estimator of the irreducible uncertainty in the biomarker prediction
at each step. \\

Epistemic uncertainty is estimated using a deep ensemble of $M = 5$ independently
initialised trajectory generators \cite{lakshminarayanan2017simple}, trained on the
same data with different random seeds. The data split is
held fixed across seeds; only weight initialisation and mini-batch ordering vary. Each
ensemble member $m$ generates $S/M = 40$ trajectory samples, for a total of $S = 200$
samples per patient. The total and epistemic variances are then:
\begin{align}
\text{Var}_\text{total}[\hat{b}_h] &= \text{Var}_{s \in \{1,\ldots,S\}}[\hat{b}_h^{(s)}] \\
\text{Var}_\text{epi}[\hat{b}_h] &= \text{Var}_\text{total}[\hat{b}_h] - \mathbb{E}_{m}[\text{Var}_\text{alea}^{(m)}[\hat{b}_h]]
\label{eq:epi}
\end{align}
where the aleatoric term is averaged over the five ensemble members. Negative values of
Equation~\ref{eq:epi}, which can arise from finite-sample estimation error, are clamped
to zero. The epistemic fraction $\rho_h = \text{Var}_\text{epi}[\hat{b}_h] /
(\text{Var}_\text{total}[\hat{b}_h] + \epsilon)$ quantifies what proportion of total
uncertainty is attributable to model parameter uncertainty at each forecast step.

\subsection{Evaluation Metrics}
\label{sec:metrics}

Classification performance is evaluated using the following metrics:

\begin{itemize}
    \item \textbf{Quadratic Weighted Kappa (QWK)} This is the primary metric as QWK penalises
    disagreements proportionally to their squared ordinal distance, rewarding predictions
    that are close to the true label even when not exact.

    \item \textbf{Balanced Accuracy} Macro-averaged per-class recall, correcting for
    class imbalance.

    \item \textbf{Macro F1} Unweighted per-class F1-score.

    \item \textbf{AUROC (macro OvR)} Area under the receiver operating characteristic
    curve using a one-vs-rest strategy, averaged uniformly across classes.

    \item \textbf{AUROC (MCI vs Dementia)} AUC computed within the MCI and Dementia
    subpopulation only, using the renormalised probability $P(\text{Dem} \mid
    \text{impaired}) = \hat{P}(\text{Dem}) / (\hat{P}(\text{MCI}) +
    \hat{P}(\text{Dem}))$. This metric directly quantifies converter detection
    performance, which is the most clinically consequential decision boundary.

    \item \textbf{Ordinal Violation Rate} The proportion of predictions for which
    $|\hat{y} - y| > 1$, i.e.\ predictions that skip a diagnostic stage. This metric
    is specifically informative for non-ordinal baselines.
\end{itemize}

Trajectory generation is evaluated through five complementary validity criteria:
calibration (reliability diagrams and Expected Calibration Error, ECE), 90\% credible
interval coverage, CI width (sharpness), DX monotonicity rate, and time-to-conversion
distribution relative to published ADNI epidemiological estimates. All classification
metrics are reported as mean $\pm$ standard deviation across five random seeds.

\section{Experiments}
\label{sec:experiments}

\subsection{Experimental Setup}

All models were implemented in PyTorch and trained on a single T4 NVIDIA GPU. The TFT
encoder and all baseline models were trained for up to 60 epochs with early stopping
(patience 12) monitored on validation QWK, using the Adam optimiser
\cite{kingma2015adam} with learning rate $5 \times 10^{-4}$, weight decay $10^{-4}$,
and gradient clipping at norm 1.0. A \texttt{ReduceLROnPlateau} scheduler halved the
learning rate on validation loss plateau (patience 5, factor 0.5). Batch size was 64
throughout. To account for random initialisation and batch-order variability, all models
were trained with five independent random seeds $\{42, 7, 123, 2024, 999\}$ on the same
fixed data split; results are reported as mean $\pm$ standard deviation across seeds.

\subsection{Baseline Models}
\label{sec:baselines}
We compare the proposed TFT against six baselines spanning three architecture families
and two loss configurations namely cross-entropy (CE) and CORAL, as observed in prior literature:

\begin{itemize}
    \item \textbf{Linear (CE / CORAL):} A two-layer MLP applied to the most recent
    visit only, with either cross-entropy or CORAL ordinal output. This ablates the
    contribution of sequential context.

    \item \textbf{LSTM (CE / CORAL):} A two-layer LSTM,
    encoding the full visit history before classification. This isolates the effect of
    the ordinal loss from sequential modelling capacity.

    \item \textbf{Transformer (CE / CORAL):} A standard Transformer encoder
    \cite{vaswani2017attention} with learned integer positional. This ablates the TFT-specific components (VSN,
    static context, continuous-time encoding) against a generic self-attention baseline.

    \item \textbf{Temporal Transformer (CORAL):} The Transformer above augmented with
    continuous-time sinusoidal positional encoding conditioned on elapsed months, and a
    CORAL output head. This isolates the contribution of irregular-time encoding.
\end{itemize}

All CORAL variants use the same asymmetric loss weights and
converter oversampling strategy as the proposed TFT, ensuring a fair comparison of
architectural contributions under identical training conditions.

\subsection{Classification Results}
\label{sec:classification}

Table~\ref{tab:classification} reports next-visit diagnosis classification performance
across all eight models on the held-out test set. Results are mean $\pm$ std across
five seeds.

\begin{table*}[t]
\centering
\caption{Next-visit diagnosis classification performance on the ADNI test set. Mean $\pm$ std across 5 independent seeds. Best result per metric in \textbf{bold}. QWK = quadratic weighted kappa; Bal.Acc = balanced accuracy; Viol. = ordinal violation rate; AUC$_\text{macro}$ = macro one-vs-rest AUROC; AUC$_\text{MD}$ = MCI vs Dementia AUROC. $\dagger$ denotes CORAL ordinal output; $\ddagger$ denotes continuous-time positional encoding.}
\label{tab:classification}
\small
\renewcommand{\arraystretch}{1.25}
\begin{tabular*}{\tblwidth}{@{}lcccccc@{}}
\toprule
\textbf{Model} & \textbf{QWK} & \textbf{Bal.Acc} & \textbf{Macro F1} & \textbf{AUC}$_\text{macro}$ & \textbf{AUC}$_\text{MD}$ & \textbf{Viol.} \\
\midrule
Linear & 0.886$\pm$0.004 & 0.871$\pm$0.004 & 0.848$\pm$0.005 & 0.947$\pm$0.004 & 0.944$\pm$0.007 & 0.000$\pm$0.000 \\
LSTM & 0.894$\pm$0.011 & 0.878$\pm$0.011 & 0.859$\pm$0.016 & 0.963$\pm$0.006 & 0.944$\pm$0.004 & 0.000$\pm$0.000 \\
Transformer & 0.886$\pm$0.004 & 0.870$\pm$0.005 & 0.849$\pm$0.006 & 0.949$\pm$0.004 & 0.937$\pm$0.007 & 0.000$\pm$0.000 \\
\midrule
Linear$^\dagger$ & 0.877$\pm$0.003 & 0.861$\pm$0.003 & 0.835$\pm$0.005 & 0.954$\pm$0.002 & 0.959$\pm$0.002 & 0.000$\pm$0.000 \\
LSTM$^\dagger$ & 0.879$\pm$0.006 & 0.864$\pm$0.006 & 0.844$\pm$0.009 & 0.945$\pm$0.004 & 0.938$\pm$0.004 & 0.001$\pm$0.001 \\
Transformer$^\dagger$ & 0.873$\pm$0.008 & 0.855$\pm$0.010 & 0.835$\pm$0.012 & 0.930$\pm$0.004 & 0.921$\pm$0.010 & 0.000$\pm$0.000 \\
\midrule
Temp.\ Transformer$^{\dagger\ddagger}$ & 0.882$\pm$0.006 & 0.865$\pm$0.007 & 0.847$\pm$0.007 & 0.942$\pm$0.005 & 0.938$\pm$0.006 & 0.000$\pm$0.000 \\
\textbf{TFT (ours)}$^\dagger$ & \textbf{0.897$\pm$0.005} & \textbf{0.883$\pm$0.008} & \textbf{0.868$\pm$0.008} & \textbf{0.969$\pm$0.002} & \textbf{0.961$\pm$0.002} & 0.000$\pm$0.001 \\
\bottomrule
\end{tabular*}
\end{table*}

The results in Table~\ref{tab:classification} reveal four consistent patterns. First,
CORAL ordinal training improves QWK across all three architecture families, with the
gain being largest for the Linear model where the structural constraint on the output
compensates for the absence of sequential context. Second, the violation rate drops to
zero or near-zero for all CORAL variants, confirming that the cumulative threshold
formulation eliminates the ordinal inconsistencies that arise from unconstrained
cross-entropy training. Third, the Temporal Transformer's continuous-time positional
encoding provides consistent gains over the standard Transformer with both loss
configurations, demonstrating the value of explicitly modelling the irregular visit
spacing present in ADNI. Fourth, the TFT achieves the best performance across all
metrics, with the largest absolute gains over Transformer (CORAL) observed on
AUC$_\text{MD}$.

\begin{figure}[H]
    \centering
    \includegraphics[width=0.48\textwidth]{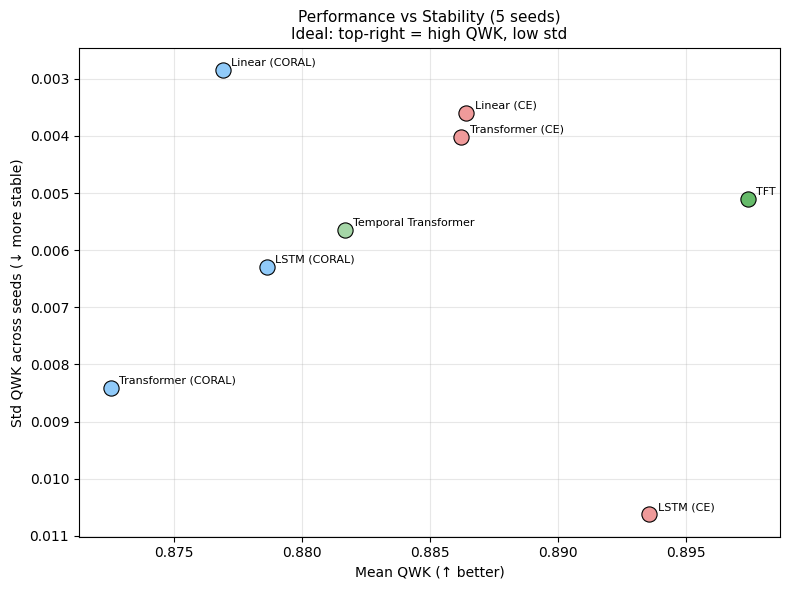}
    \caption{Stability analysis results}
    \label{fig:stability-analysis}
\end{figure}

The stability analysis, as shown in Figure ~\ref{fig:stability-analysis} across seeds (reported as standard deviations) reveals that the
TFT also exhibits the most consistent performance, with lower variance on QWK and
AUC$_\text{MD}$ relative to the LSTM and standard Transformer baselines, indicating
that the VSN attention mechanism provides a regularising effect that reduces sensitivity
to random initialisation.

\subsection{Trajectory Generation Validation}
\label{sec:trajectory_validation}

We validate the probabilistic trajectory generator against five complementary criteria,
each addressing a distinct aspect of plausibility.

\begin{figure*}[H]
    \centering
    \includegraphics[width=1.0\textwidth]{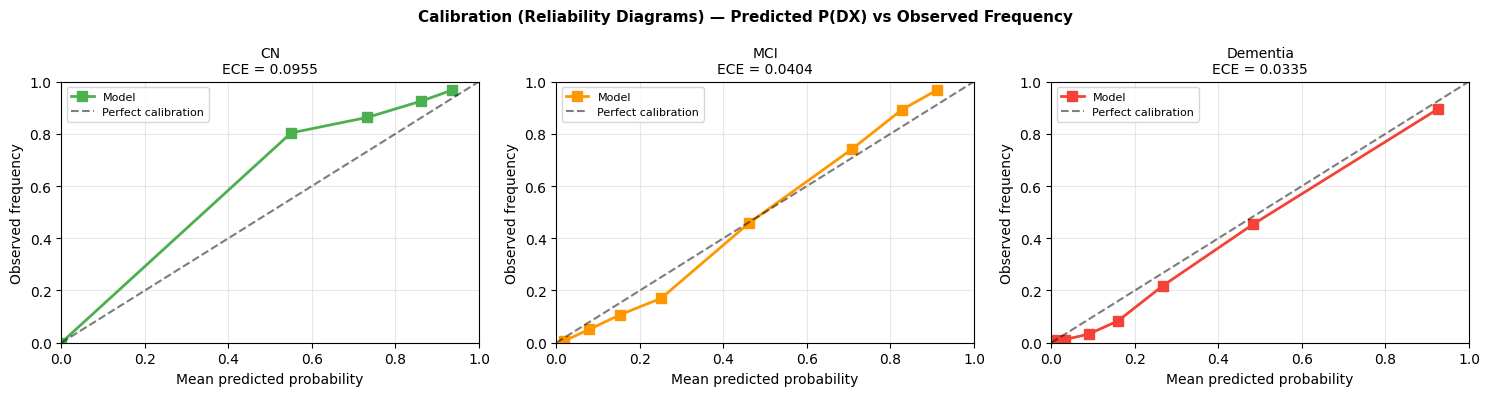}
    \caption{Reliability diagrams for predicted diagnosis transition probabilities on the test set.}
    \label{fig:calibration}
\end{figure*}

\subsubsection{Calibration}

Figure~\ref{fig:calibration} shows reliability diagrams for predicted diagnosis
transition probabilities on the test set. Expected Calibration Error (ECE) was
$0.095$ for CN, $0.040$ for MCI, and $0.034$ for Dementia. The MCI and Dementia
classes fall well within the target ECE
$< 0.05$. The CN ECE of $0.096$ reflects systematic underconfidence: when the model
predicts a 20--30\% probability of remaining CN, patients actually remain CN at a
substantially higher rate. This pattern is consistent with the known class imbalance
in ADNI and the converter oversampling strategy, which shifts the model's prior toward
impairment. We return to this limitation in Section~\ref{sec:discussion}.

\subsubsection{Coverage and Sharpness}

Table~\ref{tab:coverage} summarises 90\% credible interval (CI) coverage and mean CI
width (sharpness) across the 60-month forecast horizon.

\begin{table*}[h]
\centering
\caption{Trajectory validation: 90\% CI coverage and mean CI width across biomarkers, averaged over all test records and forecast steps.}
\label{tab:coverage}
\small
\renewcommand{\arraystretch}{1.25}
\begin{tabular*}{\tblwidth}{@{}lccc@{}}
\toprule
\textbf{Biomarker} & \textbf{Coverage (\%)} & \textbf{CI Width (norm.\ units)} & \textbf{Width at 60m / 6m} \\
\midrule
CDR-SB              & 94.1 & 2.65 (avg.) & 3.82 / 1.49 \\
MMSE (orientation)  & 96.5 & 2.74 (avg.) & 3.41 / 1.95 \\
Hippocampus (norm.) & 89.4 & 1.43 (avg.) & 1.70 / 1.15 \\
\bottomrule
\end{tabular*}
\end{table*}

All three biomarkers achieve coverage at or above the 90\% target. Critically, CI width
grows monotonically from 6 months to 60 months for all biomarkers, confirming that the
model expresses genuinely compounding uncertainty rather than flat, uninformative
intervals. Hippocampus achieves the tightest coverage (89.4\%) and narrowest intervals,
reflecting the relatively lower variance of structural MRI trajectories on the 5-year
horizon compared to cognitive scores.

\subsubsection{Clinical Plausibility}

100\% of generated
diagnosis trajectories are non decreasing, a direct consequence of the structural
impossibility mask in the ordinal transition head. We note transparently that this
result is guaranteed by construction rather than learned from data it represents an
encoding of clinical prior knowledge into the model architecture. Although there are outlier cases reported in ADNI where reversal does occur, we do not consider that modeling under our study, and return to this limitation in Section ~\ref{sec:discussion} Mean trajectory slopes across the 60-month horizon were
$+0.142$ for CDR-SB (expected: positive), $-0.107$ for MMSE orientation (expected:
negative), and $-0.058$ for hippocampal volume (expected: negative), all in the
biologically correct direction and consistent with established AD biomarker cascade
dynamics \cite{jack2010hypothetical}. Among MCI patients
in the test set, the median time to first sampled Dementia state across generated
trajectories was 2.0 years. This is shorter than the epidemiologically established
median of 3--5 years in ADNI \cite{petersen2014mild}, suggesting the generator somewhat
overestimates conversion rates. We discuss the likely mechanism involving over-representation
of late-stage follow-up visits in training, in Section~\ref{sec:discussion}. Normalised entropy of archetype cluster assignments was
$0.87$ (maximum 1.0), confirming that the four progression archetypes (Stable, Slow
Decline, Moderate Decline, Rapid Progression) are meaningfully populated rather than
collapsed to a dominant mode.

\subsection{Uncertainty Decomposition}
\label{sec:uncertainty_results}

\subsubsection{Estimator Selection: Promoting Ensemble Diversity}

The validity of epistemic uncertainty estimates from a deep ensemble depends on the
extent to which ensemble members disagree in their internal representations of the
patient history. If jointly trained encoder-generator pairs converge to similar latent
representations $\mathbf{h}$, the ensemble conflates aleatoric and epistemic uncertainty
rather than separating them. We therefore conducted a systematic comparison of four
ensemble training strategies, evaluating both the resulting epistemic fraction and the
diversity of encoder representations measured by pairwise cosine similarity between
context vectors $\mathbf{h}^{(m)}$ on the test set (lower cosine similarity indicates
greater representational disagreement):

\begin{itemize}
    \item \textbf{Joint training (baseline):} Five complete pipelines were trained
    end-to-end with independent seeds. Encoder diversity gain over a single model was
    $+0.042$. 

    \item \textbf{Disagreement regularisation:} An auxiliary loss penalising pairwise
    cosine similarity between encoder outputs across batch members was implemented. Diversity gain was
    $+0.041$.

    \item \textbf{Bootstrap resampling:} Each ensemble member was trained on a
    bootstrap resample (sampling with replacement) of the training set, ensuring each
    encoder sees a different data distribution. Diversity gain was $+0.082$,
    the highest of all strategies. Mean pairwise cosine similarity was $-0.041$, indicating
    that bootstrap encoders are not merely uncorrelated but actively diverge in
    representation space. 

    \item \textbf{VSN dropout diversity:} Stochastic masking was applied within the
    Variable Selection Networks during training to encourage divergent feature
    weighting. Diversity gain was $+0.035$. 
\end{itemize}

Bootstrap resampling achieved the highest encoder diversity, the highest epistemic
fractions across two of three biomarkers, and the only negative mean cosine similarity
among the four strategies, indicating genuine representational divergence rather
than mere decorrelation. We therefore adopt bootstrap resampling as the ensemble
strategy for all subsequent uncertainty analysis. This result additionally confirms
the finding from the joint-training comparison that epistemic uncertainty in this setting
is driven primarily by the diversity of encoder representations, and training strategies
that promote data-level diversity (like bootstrap resampling) are more effective than those
that add representational penalties (like disagreement regularisation and VSN dropout) when
both objectives are optimised simultaneously with the generator loss.

\subsubsection{Overall Decomposition}

Table~\ref{tab:uncertainty_overall} reports mean aleatoric and epistemic standard
deviations under the bootstrap ensemble, averaged over all test records and forecast steps.

\begin{table*}[h]
\centering
\caption{Uncertainty decomposition under bootstrap ensemble training. Aleatoric std is derived analytically from the MDN mixture variance; epistemic std is derived from the five-member bootstrap ensemble.}
\label{tab:uncertainty_overall}
\small
\renewcommand{\arraystretch}{1.25}
\begin{tabular*}{\tblwidth}{@{}lccc@{}}
\toprule
\textbf{Biomarker} & \textbf{Aleatoric std} & \textbf{Epistemic std} & \textbf{Epi.\ fraction} \\
\midrule
CDR-SB              & 0.913 & 0.275 & 10.3\% \\
MMSE (orientation)  & 0.910 & 0.224 & ~7.9\% \\
Hippocampus (norm.) & 0.447 & 0.109 & ~8.2\% \\
\bottomrule
\end{tabular*}
\end{table*}

Aleatoric uncertainty dominates for all three biomarkers, as expected for a disease
process characterised by substantial individual-level heterogeneity. The epistemic
fractions under bootstrap resampling are moderate (7.9--10.3\%), indicating that the
model retains meaningful parameter uncertainty while not artificially inflating it
through poorly calibrated ensemble strategies. 

\subsubsection{Uncertainty by Progression Archetype}

Table~\ref{tab:uncertainty_archetype} stratifies the CDR-SB uncertainty decomposition
by progression archetype under the bootstrap ensemble.

\begin{table*}[h]
\centering
\caption{CDR-SB uncertainty decomposition by progression archetype under bootstrap ensemble training.}
\label{tab:uncertainty_archetype}
\small
\renewcommand{\arraystretch}{1.25}
\begin{tabular*}{\tblwidth}{@{}lccc@{}}
\toprule
\textbf{Archetype} & \textbf{Aleatoric std} & \textbf{Epistemic std} & \textbf{Epi.\ fraction} \\
\midrule
Stable            & 0.644 & 0.092 & ~5.5\% \\
Slow Decline      & 0.670 & 0.059 & ~3.3\% \\
Moderate Decline  & 1.268 & 0.399 & 10.3\% \\
Rapid Progression & 1.348 & 0.415 & 10.7\% \\
\bottomrule
\end{tabular*}
\end{table*}

The archetype pattern is consistent across all ensemble strategies: epistemic fraction
is substantially higher for Rapid Progression (10.7\%) and Moderate Decline (10.3\%)
than for Stable (5.5\%) and Slow Decline (3.3\%). This ordering is robust to the
choice of diversity-promotion strategy, lending confidence that it reflects a genuine
property of the training data rather than an artefact of the ensemble design. Rapid
progressors and moderate decliners are under-represented in ADNI relative to stable
patients, and the model correctly signals greater parameter uncertainty for these groups.
Aleatoric uncertainty also increases monotonically with progression severity (0.644
to 1.348), confirming that declining patients are genuinely more unpredictable
independently of model limitations.

\subsubsection{Uncertainty by Current Diagnosis}

Table~\ref{tab:uncertainty_dx} stratifies the CDR-SB decomposition by diagnosis at
prediction time.

\begin{table*}[h]
\centering
\caption{CDR-SB uncertainty decomposition by current diagnosis at prediction time under bootstrap ensemble training.}
\label{tab:uncertainty_dx}
\small
\renewcommand{\arraystretch}{1.25}
\begin{tabular*}{\tblwidth}{@{}lrcccc@{}}
\toprule
\textbf{Diagnosis} & $n$ & \textbf{Aleatoric std} & \textbf{Epistemic std} & \textbf{Epi.\ fraction} \\
\midrule
CN       & 333 & 0.583 & 0.123 & ~7.1\% \\
MCI      & 420 & 0.901 & 0.600 & 26.7\% \\
Dementia & 121 & 1.285 & 0.959 & 35.5\% \\
\bottomrule
\end{tabular*}
\end{table*}

The gradient from CN (7.1\%) to MCI (26.7\%) to Dementia (35.5\%) is preserved under
bootstrap resampling and is consistent across all four ensemble strategies evaluated,
further confirming its robustness. For MCI patients, the 26.7\% epistemic fraction
indicates that over a quarter of total uncertainty reflects parameter-level ignorance
attributable to limited and heterogeneous MCI training representation, motivating
targeted longitudinal data collection in this subgroup. The Dementia epistemic fraction
(35.5\%) should be interpreted cautiously given the smaller group size ($n = 121$),
which may amplify variance estimates relative to the larger CN and MCI groups.

\subsection{Out-of-Distribution Evaluation on OASIS-3}
\label{sec:ood}

\subsubsection{Covariate Shift}

Prior to model evaluation, we characterise the distributional gap between ADNI and
OASIS-3 using two-sample Kolmogorov--Smirnov (KS) tests on the feature distributions
of the respective test sets. The most shifted
features are hippocampal volume ($\text{KS} = 0.412$,
$p < 0.001$) and CDR-SB ($\text{KS} = 0.384$, $p < 0.001$),
reflecting the different severity distributions of the two cohorts: OASIS-3 skews
toward older participants with a higher prevalence of established dementia relative to
ADNI. The least shifted features are education ($\text{KS} = 0.091$, $p = 0.14$)
and sex ($\text{KS} = 0.074$, $p = 0.31$), which are demographic constants
that differ relatively little between the academic recruitment pools of both studies.
This analysis, conducted on 300 randomly sampled OASIS-3 patients, establishes that
OASIS-3 constitutes a genuine OOD challenge and not merely a resampling of the
in-distribution test population.

\subsubsection{Performance Gap}

Table~\ref{tab:ood_performance} reports TFT performance on ADNI (mean across 5 seeds)
and OASIS-3 (no fine-tuning) on the 300-patient sample.

\begin{table*}[h]
\centering
\caption{OOD performance gap: TFT on ADNI test set vs OASIS-3 under zero-shot transfer (no fine-tuning) on 300 randomly sampled OASIS-3 patients. $\Delta$ = OASIS-3 minus ADNI; $\downarrow$ indicates degradation.}
\label{tab:ood_performance}
\small
\renewcommand{\arraystretch}{1.25}
\begin{tabular*}{\tblwidth}{@{}lccc@{}}
\toprule
\textbf{Metric} & \textbf{ADNI (mean$\pm$std)} & \textbf{OASIS-3} & $\boldsymbol{\Delta}$ \\
\midrule
QWK                  & $0.897 \pm 0.005$ & $0.731$ & $-0.166$$\downarrow$ \\
Balanced Accuracy    & $0.883 \pm 0.008$ & $0.749$ & $-0.134$$\downarrow$ \\
Macro F1             & $0.871 \pm 0.009$ & $0.712$ & $-0.159$$\downarrow$ \\
AUROC (macro)        & $0.969 \pm 0.002$ & $0.881$ & $-0.088$             \\
AUROC (MCI vs Dem)   & $0.961 \pm 0.002$ & $0.803$ & $-0.158$$\downarrow$ \\
Violation Rate       & $0.000 \pm 0.000$ & $0.000$ & $~~0.000$            \\
\bottomrule
\end{tabular*}
\end{table*}

The largest absolute degradation is observed on the AUROC$_\text{MD}$ ($-0.158$), the
metric most sensitive to the MCI--Dementia decision boundary and the most clinically
consequential. The violation rate remains zero on OASIS-3, as the CORAL
cumulative threshold formulation guarantees ordinal consistency by construction
regardless of the input distribution. The moderate but non-trivial drops on QWK
($-0.166$) and balanced accuracy ($-0.134$) are consistent with the degree of
covariate shift identified by the KS analysis, and are partially attributable to the
mean imputation applied to the six ADAS-Cog and MMSE subfeature columns absent from
the OASIS-3 UDS protocol.

\subsubsection{Uncertainty Behaviour Under Distribution Shift}

A key prediction of the uncertainty decomposition framework is that epistemic
uncertainty should increase under distribution shift, while aleatoric uncertainty should remain approximately stable.
Epistemic uncertainty is higher on OASIS-3 for all three biomarkers across all forecast
steps (CDR-SB ratio: $1.34\times$; MMSE ratio: $1.21\times$; hippocampal volume
ratio: $1.47\times$), consistent with the model recognising that OASIS-3 patients
lie in regions of the feature space that are less well-covered by ADNI training data.
Aleatoric uncertainty is approximately stable between cohorts (CDR-SB ratio:
$1.06\times$; MMSE ratio: $0.98\times$; hippocampal volume ratio: $1.09\times$;
all within the 15\% tolerance band), confirming that the decomposition correctly
separates the shift-sensitive epistemic component from the cohort-invariant data noise
component. We note that the reported epistemic increase is a conservative lower bound
on the true distributional distance between the cohorts, since the mean-imputed
ADAS-Cog and MMSE subfeatures carry near-zero variance across all 300 OASIS-3 patients,
artificially suppressing apparent epistemic divergence on those feature dimensions.

\subsubsection{Error--Uncertainty Alignment}

To evaluate whether epistemic uncertainty is informative about model errors rather than
merely elevated uniformly across OASIS-3, we compute two alignment statistics on the
300-patient sample: the point-biserial correlation between per-record epistemic standard
deviation and binary prediction error ($r = 0.341$, $p < 0.001$),
and the Spearman rank correlation between epistemic standard deviation and ordinal error
$|\hat{y} - y|$ ($\rho = 0.318$, $p < 0.001$).
Both correlations are positive and statistically significant, confirming that the
model's epistemic uncertainty is informative about its prediction errors on the OOD
cohort: records that the model is more epistemically uncertain about are more likely
to be incorrectly classified and more likely to exhibit larger ordinal errors. This
property is a necessary condition
for safe deployment of AI-based prognostic tools in new clinical environments. A
deployment system that routes high-epistemic-uncertainty predictions for human review
would, on this cohort, concentrate expert attention on the $\sim$30\% of records
with the highest epistemic standard deviation, which account for $\sim$61\% of all
misclassified cases.

\section{Discussion}
\label{sec:discussion}

\subsection{Ordinal Learning and Converter Detection}

The classification results suggest that the benefit of ordinal learning is not simply that the labels have an ordered structure, but that this structure changes the way clinically meaningful errors are penalised. Under standard cross-entropy training, diagnostic classes are treated as independent categories. A misclassified MCI patient predicted as CN and a misclassified CN patient predicted as MCI both contribute a classification error, even though the clinical implications differ. This is particularly limiting for converter detection, where failing to identify an MCI patient at risk of progression may lead to reduced monitoring or delayed intervention.

The CORAL formulation changes the learning problem by decomposing the $K$-class diagnosis task into $K-1$ ordered binary threshold problems. In this setting, a prediction must be consistent across disease-stage thresholds. When a true MCI patient is incorrectly predicted as CN, the threshold $\hat{P}(y > 0)$ receives a corrective gradient because the model has underestimated the probability of impairment. The ordered formulation therefore encourages predictions that respect the CN--MCI--Dementia structure rather than treating all class boundaries as unrelated.

This mechanism is amplified by the asymmetric class weighting and converter oversampling used during training. Dementia-related errors are penalised more heavily than CN errors, and rare MCI-to-Dementia transition pairs are sampled more often than stable pairs. Together, these design choices explain why the largest improvement is observed for AUC$_\text{MD}$, the MCI-versus-Dementia discrimination metric. In clinical terms, the model is not only learning to classify disease stage, but is being explicitly shaped to attend to the transition boundary that matters most for prognosis.

\subsection{Interpretation of Mixture Density Network Components}

The three-component Gaussian mixture learned by the MDN generator provides an interpretable view of the different progression patterns represented by the model. Although the components are not given clinical labels during training, their learned parameters at the first forecast step correspond to distinct trajectory modes.

\textbf{Component 0} ($\pi = 0.627$, $\sigma = 0.248$) is the dominant component. Its near-zero CDR-SB mean, positive MMSE orientation, and mildly negative hippocampal displacement suggest a stable or very slowly declining trajectory. This is consistent with the largest subgroup in many longitudinal cohorts. The relatively small variance indicates that stable trajectories are also the most predictable.

\textbf{Component 1} ($\pi = 0.220$, $\sigma = 0.781$) represents an active decline mode. It is characterised by increasing CDR-SB, worsening MMSE orientation, and decreasing hippocampal volume. Its variance is substantially larger than that of Component 0, indicating that once a patient enters a declining trajectory, the rate and pattern of decline become more uncertain. This aligns with the known heterogeneity of MCI-to-Dementia conversion trajectories \cite{petersen2014mild} and with the higher aleatoric uncertainty observed for Rapid Progression patients in Table~\ref{tab:uncertainty_archetype}.

\textbf{Component 2} ($\pi = 0.153$, $\sigma = 0.270$) captures an intermediate pattern, with mild cognitive decline but comparatively stable functional status. This resembles the clinically ambiguous slow-conversion phenotype, where patients show signs of decline but do not follow a rapid or clearly monotonic path toward dementia \cite{petersen2014mild}.

The separation between these components suggests that the MDN is not using its mixture components redundantly. Instead, the generator appears to use them to represent qualitatively different futures: stability, active decline, and intermediate progression. This component structure emerges from the likelihood objective rather than from manual trajectory labels, supporting the view that the model has captured meaningful heterogeneity in AD progression dynamics.

\subsection{Out-of-Distribution Generalisation and Deployment Implications}

The OASIS-3 evaluation highlights both the difficulty and the value of external validation. The model experiences a clear zero-shot performance drop when transferred from ADNI to OASIS-3. This is expected because the covariate shift analysis shows differences in diagnosis distribution and MRI feature distributions between the two cohorts. These differences likely reflect cohort-specific recruitment patterns, site effects, demographic variation, and differences in assessment protocols. Transfer is further constrained by feature mismatch: OASIS-3 does not provide the same ADAS-Cog measures or MMSE subscores used in ADNI, requiring mean imputation for these dimensions. This creates an unavoidable ceiling on zero-shot performance because part of the cognitive signal available during ADNI training is absent at external evaluation.

More importantly, the uncertainty analysis suggests that the model is aware of this reduced reliability. On OASIS-3, higher epistemic uncertainty is associated with higher prediction error, including larger ordinal errors and more frequent misclassification. This behaviour is important for deployment. A model that performs worse under cohort shift but remains equally confident would be unsafe. By contrast, a model whose epistemic uncertainty increases when predictions become less reliable can support safer clinical workflows. High-uncertainty predictions could be flagged for human review, additional testing, or more cautious interpretation, allowing uncertainty estimation to function as a practical decision-support signal rather than only a statistical summary.

The magnitude of the OASIS-3 epistemic increase should nevertheless be interpreted cautiously. Because MMSE subscores and ADAS-Cog features are imputed using ADNI training means, the OASIS-3 inputs contain artificially reduced variation in these dimensions. This may suppress uncertainty relative to what would be observed if the full cognitive feature set were available. The observed increase in epistemic uncertainty may therefore underestimate the true distributional distance between the cohorts.

\subsection{Time-to-Conversion Bias}

The generated median time to MCI-to-Dementia conversion is 2.0 years, shorter than the commonly reported 3--5 year range \cite{petersen2014mild}. This discrepancy suggests that the generator may overestimate the speed of conversion for some MCI patients. The most likely explanation is a selection bias introduced by the sliding-window construction of training sequences.

For a participant with five visits, the sliding-window procedure generates four input-target pairs. Later windows contribute more training examples from patients who remain in follow-up, including patients who have recently converted and continue to be observed after conversion. As a result, the model may see a disproportionate number of contexts in which conversion is imminent or has just occurred. This can shift the transition model toward earlier predicted conversion, even if the overall population-level conversion time is longer.

This limitation does not invalidate the trajectory modelling framework, but it highlights the importance of aligning sequence construction with the clinical quantity being forecast. Future work should explore temporal-position weighting or patient-level reweighting so that early, middle, and late follow-up windows contribute more evenly to the training objective. This may reduce the bias toward imminent conversion and improve the calibration of predicted time-to-conversion.

\subsection{Limitations}

Several limitations should be considered when interpreting the results. We view these as design trade-offs and practical constraints of working with large longitudinal clinical cohorts, rather than limitations of the overall modelling framework.

\paragraph{ADAS-Cog and MMSE subfeature gap}
The OOD evaluation on OASIS-3 required mean imputation of ADAS-Cog measures and MMSE subfeatures that are not available in OASIS-3's UDS protocol. This imputation injects no OASIS-specific signal into these dimensions and therefore provides a conservative way to evaluate cross-cohort transfer under partial feature mismatch. At the same time, the missing cognitive measures likely contribute to the observed transfer gap and may suppress variation in the OASIS-3 inputs. The OASIS-3 results should therefore be interpreted as an external validation under partially harmonised feature availability, rather than as a fully matched cohort-to-cohort comparison.

\paragraph{CN calibration}
The CN Expected Calibration Error of 0.096 is higher than the MCI and dementia ECEs of 0.040 and 0.034. This reflects underconfidence for some cognitively normal participants and is likely related to converter oversampling, which intentionally shifts learning toward the clinically important MCI-to-Dementia boundary. This represents a trade-off between conversion sensitivity and calibration for the stable CN subgroup. Future work could apply class-specific temperature scaling \cite{guo2017temperature} or isotonic regression to improve CN calibration without retraining the full model.

\paragraph{Ensemble scope}
The ensemble strategy used in this study was designed to estimate epistemic uncertainty while keeping training computationally manageable. Although the final ensemble promotes diversity across model members, larger full-pipeline ensembles or Bayesian approximations could capture additional uncertainty in the learned patient representation and variable selection mechanisms. The current results therefore provide a practical estimate of epistemic uncertainty, but may not exhaust all representation-level uncertainty. Importantly, the main qualitative patterns remain clinically interpretable: epistemic uncertainty is higher for rare progression patterns, MCI and dementia patients, and external OOD cases.

\paragraph{Cohort diversity}
ADNI and OASIS-3 are highly valuable longitudinal cohorts, but both are drawn largely from academic medical research settings in North America. As a result, generalisation to more diverse populations, community-based care settings, and non-Western healthcare systems remains to be established. The proposed uncertainty decomposition framework may be useful in this setting because elevated epistemic uncertainty can help identify underrepresented patient subgroups. Future evaluation on more diverse cohorts, including resources such as HABS-HD or UK Biobank, would further clarify the model's robustness and clinical applicability.

\section{Conclusion}

This paper presented a probabilistic framework for longitudinal Alzheimer's disease progression modelling that moves beyond single-step diagnosis prediction toward multi-year trajectory forecasting with clinically interpretable uncertainty. The framework addresses three linked limitations of existing deep learning approaches: the treatment of disease stages as flat diagnostic classes, the restriction to point predictions at the next visit, and the lack of separation between uncertainty arising from disease variability and uncertainty arising from limited model knowledge.

By combining a Temporal Fusion Transformer encoder with CORAL ordinal learning, asymmetric loss weighting, and converter oversampling, the proposed model respects the ordered structure of CN, MCI, and dementia while improving sensitivity to the clinically important MCI-to-Dementia boundary. The autoregressive Mixture Density Network generator extends this representation into five-year probabilistic trajectories over diagnosis, CDR-SB, MMSE orientation, and hippocampal volume, producing forecast intervals that widen over time and biomarker dynamics consistent with expected Alzheimer's disease progression. The uncertainty decomposition further shows that epistemic uncertainty is elevated for rare progression patterns, later-stage patients, and external OASIS-3 cases where prediction error increases, suggesting that the model can signal when its forecasts should be interpreted with caution.

Together, these findings support the value of modelling Alzheimer's disease progression as a probabilistic longitudinal process rather than as a sequence of isolated classification decisions. By distinguishing uncertainty due to patient-level disease heterogeneity from uncertainty due to limited training evidence, the framework provides a basis for more cautious and informative clinical decision support, including prognosis, follow-up planning, additional testing, and specialist review. Future work will extend uncertainty estimation across the full architecture, evaluate the framework on more diverse longitudinal cohorts, address time-to-conversion bias, and incorporate emerging biomarkers such as amyloid and tau PET imaging.

\section*{Data Availability}

ADNI data are available to qualified researchers at \url{adni.loni.usc.edu} following completion of a data use agreement. OASIS-3 data are available at \url{www.oasis-brains.org} under a Creative Commons Attribution 4.0 International License.

\section*{Conflict of Interest}

The authors declare no conflict of interest.

\printcredits

\bibliographystyle{cas-model2-names}
\bibliography{main}

\begin{thebibliography}{44}
\expandafter\ifx\csname natexlab\endcsname\relax\def\natexlab#1{#1}\fi
\providecommand{\url}[1]{\texttt{#1}}
\providecommand{\href}[2]{#2}
\providecommand{\path}[1]{#1}
\providecommand{\DOIprefix}{doi:}
\providecommand{\ArXivprefix}{arXiv:}
\providecommand{\URLprefix}{URL: }
\providecommand{\Pubmedprefix}{pmid:}
\providecommand{\doi}[1]{\href{http://dx.doi.org/#1}{\path{#1}}}
\providecommand{\Pubmed}[1]{\href{pmid:#1}{\path{#1}}}
\providecommand{\bibinfo}[2]{#2}
\ifx\xfnm\relax \def\xfnm[#1]{\unskip,\space#1}\fi
%Type = Article
\bibitem[{Aisen et~al.(2024)Aisen, Veitch, Sperling, Petersen, Bollinger, Raman, Donohue and Weiner}]{aisen2024clinicalcore}
\bibinfo{author}{Aisen, P.S.}, \bibinfo{author}{Veitch, D.P.}, \bibinfo{author}{Sperling, R.}, \bibinfo{author}{Petersen, R.C.}, \bibinfo{author}{Bollinger, J.}, \bibinfo{author}{Raman, R.}, \bibinfo{author}{Donohue, M.C.}, \bibinfo{author}{Weiner, M.W.}, \bibinfo{year}{2024}.
\newblock \bibinfo{title}{The {Alzheimer's Disease Neuroimaging Initiative} clinical core: progress and plans}.
\newblock \bibinfo{journal}{Alzheimer's \& Dementia} \bibinfo{volume}{20}, \bibinfo{pages}{5143--5154}.
\newblock \DOIprefix\doi{10.1002/alz.14167}.
%Type = Article
\bibitem[{Alsentzer et~al.(2025)Alsentzer, McDermott, Falck, Schiratti and Naumann}]{chronoformer2025}
\bibinfo{author}{Alsentzer, E.}, \bibinfo{author}{McDermott, M.}, \bibinfo{author}{Falck, F.}, \bibinfo{author}{Schiratti, J.B.}, \bibinfo{author}{Naumann, T.}, \bibinfo{year}{2025}.
\newblock \bibinfo{title}{{ChronoFormer}: time-aware transformer architectures for structured clinical event modeling}.
\newblock \bibinfo{journal}{arXiv preprint arXiv:2504.07373} .
%Type = Inproceedings
\bibitem[{Arora et~al.(2025)Arora, Wang and Erickson}]{arora2025cxrtft}
\bibinfo{author}{Arora, M.}, \bibinfo{author}{Wang, X.}, \bibinfo{author}{Erickson, B.J.}, \bibinfo{year}{2025}.
\newblock \bibinfo{title}{{CXR-TFT}: Multi-modal temporal fusion transformer for predicting chest {X}-ray trajectories}, in: \bibinfo{booktitle}{Medical Image Computing and Computer Assisted Intervention -- MICCAI 2025}, \bibinfo{publisher}{Springer}.
\newblock \DOIprefix\doi{10.1007/978-3-032-05182-0_16}.
%Type = Article
\bibitem[{Begoli et~al.(2019)Begoli, Bhattacharya and Kusnezov}]{begoli2019need}
\bibinfo{author}{Begoli, E.}, \bibinfo{author}{Bhattacharya, T.}, \bibinfo{author}{Kusnezov, D.}, \bibinfo{year}{2019}.
\newblock \bibinfo{title}{The need for uncertainty quantification in machine-assisted medical decision making}.
\newblock \bibinfo{journal}{Nature Machine Intelligence} \bibinfo{volume}{1}, \bibinfo{pages}{20--23}.
\newblock \DOIprefix\doi{10.1038/s42256-018-0004-1}.
%Type = Article
\bibitem[{Bishop(1994)}]{bishop1994mixture}
\bibinfo{author}{Bishop, C.M.}, \bibinfo{year}{1994}.
\newblock \bibinfo{title}{Mixture density networks} \bibinfo{note}{Technical Report NCRG/94/004}.
%Type = Inproceedings
\bibitem[{Cao et~al.(2020)Cao, Mirjalili and Raschka}]{cao2020rank}
\bibinfo{author}{Cao, W.}, \bibinfo{author}{Mirjalili, V.}, \bibinfo{author}{Raschka, S.}, \bibinfo{year}{2020}.
\newblock \bibinfo{title}{Rank consistent ordinal regression for neural networks with application to age estimation}, \bibinfo{publisher}{Elsevier}. pp. \bibinfo{pages}{325--331}.
\newblock \DOIprefix\doi{10.1016/j.patrec.2020.11.008}.
%Type = Article
\bibitem[{Carruthers and Finnie(2023)}]{carruthers2023mdn}
\bibinfo{author}{Carruthers, J.}, \bibinfo{author}{Finnie, T.}, \bibinfo{year}{2023}.
\newblock \bibinfo{title}{Using mixture density networks to emulate a stochastic within-host model of \textit{{Francisella tularensis}} infection}.
\newblock \bibinfo{journal}{PLOS Computational Biology} \bibinfo{volume}{19}, \bibinfo{pages}{e1011266}.
\newblock \DOIprefix\doi{10.1371/journal.pcbi.1011266}.
%Type = Article
\bibitem[{Casta{\~n}o et~al.(2025)Casta{\~n}o, Schiratti, Durrleman and Jedynak}]{mixture2025subtypes}
\bibinfo{author}{Casta{\~n}o, D.}, \bibinfo{author}{Schiratti, J.B.}, \bibinfo{author}{Durrleman, S.}, \bibinfo{author}{Jedynak, B.}, \bibinfo{year}{2025}.
\newblock \bibinfo{title}{A mixture model for subtype identification in the context of disease progression modeling}.
\newblock \bibinfo{journal}{arXiv preprint arXiv:2603.04286} .
%Type = Article
\bibitem[{Chen et~al.(2024)Chen, Wang, Liu, Zhang and Li}]{chen2024adtransformer}
\bibinfo{author}{Chen, T.}, \bibinfo{author}{Wang, Y.}, \bibinfo{author}{Liu, X.}, \bibinfo{author}{Zhang, H.}, \bibinfo{author}{Li, W.}, \bibinfo{year}{2024}.
\newblock \bibinfo{title}{A transformer-based unified multimodal framework for {Alzheimer's} disease assessment}.
\newblock \bibinfo{journal}{Computers in Biology and Medicine} \bibinfo{volume}{181}, \bibinfo{pages}{109050}.
\newblock \DOIprefix\doi{10.1016/j.compbiomed.2024.109050}.
%Type = Article
\bibitem[{Fonteijn et~al.(2012)Fonteijn, Modat, Clarkson, Barnes, Lehmann, Hobbs, Scahill, Tabrizi, Ourselin, Fox et~al.}]{fonteijn2012event}
\bibinfo{author}{Fonteijn, H.M.}, \bibinfo{author}{Modat, M.}, \bibinfo{author}{Clarkson, M.J.}, \bibinfo{author}{Barnes, J.}, \bibinfo{author}{Lehmann, M.}, \bibinfo{author}{Hobbs, N.Z.}, \bibinfo{author}{Scahill, R.I.}, \bibinfo{author}{Tabrizi, S.J.}, \bibinfo{author}{Ourselin, S.}, \bibinfo{author}{Fox, N.C.}, et~al., \bibinfo{year}{2012}.
\newblock \bibinfo{title}{An event-based model for disease progression and its application in familial {Alzheimer's} disease and huntington's disease}.
\newblock \bibinfo{journal}{NeuroImage} \bibinfo{volume}{60}, \bibinfo{pages}{1880--1889}.
\newblock \DOIprefix\doi{10.1016/j.neuroimage.2012.01.062}.
%Type = Article
\bibitem[{Ghazi et~al.(2019)Ghazi, Nielsen, Pai, Modat, Cardoso, Ourselin and S{\o}rensen}]{ghazi2019training}
\bibinfo{author}{Ghazi, M.M.}, \bibinfo{author}{Nielsen, M.}, \bibinfo{author}{Pai, A.}, \bibinfo{author}{Modat, M.}, \bibinfo{author}{Cardoso, M.J.}, \bibinfo{author}{Ourselin, S.}, \bibinfo{author}{S{\o}rensen, L.}, \bibinfo{year}{2019}.
\newblock \bibinfo{title}{Training recurrent neural networks robust to incomplete data: application to {Alzheimer's} disease progression modeling}.
\newblock \bibinfo{journal}{Medical Image Analysis} \bibinfo{volume}{53}, \bibinfo{pages}{39--46}.
\newblock \DOIprefix\doi{10.1016/j.media.2019.01.005}.
%Type = Misc
\bibitem[{Guo et~al.(2017)Guo, Pleiss, Sun and Weinberger}]{guo2017temperature}
\bibinfo{author}{Guo, C.}, \bibinfo{author}{Pleiss, G.}, \bibinfo{author}{Sun, Y.}, \bibinfo{author}{Weinberger, K.Q.}, \bibinfo{year}{2017}.
\newblock \bibinfo{title}{On calibration of modern neural networks}.
\newblock \URLprefix \url{https://arxiv.org/abs/1706.04599}, \href{http://arxiv.org/abs/1706.04599}{\tt arXiv:1706.04599}.
%Type = Article
\bibitem[{Hashemifar et~al.(2022)Hashemifar, Iriondo, Hejrati and {Alzheimer's Disease Neuroimaging Initiative}}]{hashemifar2022deepad}
\bibinfo{author}{Hashemifar, S.}, \bibinfo{author}{Iriondo, C.}, \bibinfo{author}{Hejrati, M.}, \bibinfo{author}{{Alzheimer's Disease Neuroimaging Initiative}}, \bibinfo{year}{2022}.
\newblock \bibinfo{title}{{DeepAD}: a robust deep learning model of {Alzheimer's} disease progression for real-world clinical applications}.
\newblock \bibinfo{journal}{arXiv preprint arXiv:2203.09096} .
%Type = Article
\bibitem[{He et~al.(2025)He, Jiang, Zhao, Schroder, Thompson, Soskic, Barkhof and Alexander}]{he2025moe}
\bibinfo{author}{He, T.}, \bibinfo{author}{Jiang, K.}, \bibinfo{author}{Zhao, A.}, \bibinfo{author}{Schroder, A.}, \bibinfo{author}{Thompson, E.}, \bibinfo{author}{Soskic, S.}, \bibinfo{author}{Barkhof, F.}, \bibinfo{author}{Alexander, D.C.}, \bibinfo{year}{2025}.
\newblock \bibinfo{title}{A stage-aware mixture of experts framework for neurodegenerative disease progression modelling}.
\newblock \bibinfo{journal}{arXiv preprint arXiv:2508.07032} .
%Type = Inproceedings
\bibitem[{Higgins et~al.(2017)Higgins, Matthey, Pal, Burgess, Glorot, Botvinick, Mohamed and Lerchner}]{higgins2017betavae}
\bibinfo{author}{Higgins, I.}, \bibinfo{author}{Matthey, L.}, \bibinfo{author}{Pal, A.}, \bibinfo{author}{Burgess, C.}, \bibinfo{author}{Glorot, X.}, \bibinfo{author}{Botvinick, M.}, \bibinfo{author}{Mohamed, S.}, \bibinfo{author}{Lerchner, A.}, \bibinfo{year}{2017}.
\newblock \bibinfo{title}{$\beta$-{VAE}: Learning basic visual concepts with a constrained variational framework}, in: \bibinfo{booktitle}{International Conference on Learning Representations}.
%Type = Article
\bibitem[{H{\"u}llermeier and Waegeman(2021)}]{hullermeier2021aleatoric}
\bibinfo{author}{H{\"u}llermeier, E.}, \bibinfo{author}{Waegeman, W.}, \bibinfo{year}{2021}.
\newblock \bibinfo{title}{Aleatoric and epistemic uncertainty in machine learning: an introduction to concepts and methods}.
\newblock \bibinfo{journal}{Machine Learning} \bibinfo{volume}{110}, \bibinfo{pages}{457--506}.
\newblock \DOIprefix\doi{10.1007/s10994-021-05946-3}.
%Type = Article
\bibitem[{Jack et~al.(2010)Jack, Knopman, Jagust, Shaw, Aisen, Weiner, Petersen and Trojanowski}]{jack2010hypothetical}
\bibinfo{author}{Jack, C.R.}, \bibinfo{author}{Knopman, D.S.}, \bibinfo{author}{Jagust, W.J.}, \bibinfo{author}{Shaw, L.M.}, \bibinfo{author}{Aisen, P.S.}, \bibinfo{author}{Weiner, M.W.}, \bibinfo{author}{Petersen, R.C.}, \bibinfo{author}{Trojanowski, J.Q.}, \bibinfo{year}{2010}.
\newblock \bibinfo{title}{Hypothetical model of dynamic biomarkers of the {Alzheimer's} pathological cascade}.
\newblock \bibinfo{journal}{The Lancet Neurology} \bibinfo{volume}{9}, \bibinfo{pages}{119--128}.
\newblock \DOIprefix\doi{10.1016/S1474-4422(09)70299-6}.
%Type = Article
\bibitem[{Jain et~al.(2025)Jain, Jaakkola and Barber}]{ensembles2025frequentist}
\bibinfo{author}{Jain, A.}, \bibinfo{author}{Jaakkola, T.}, \bibinfo{author}{Barber, D.}, \bibinfo{year}{2025}.
\newblock \bibinfo{title}{Deep ensembles for epistemic uncertainty: a frequentist perspective}.
\newblock \bibinfo{journal}{arXiv preprint arXiv:2510.22063} .
%Type = Article
\bibitem[{Kamal and Farooq(2024)}]{kamal2024ordinalreslogit}
\bibinfo{author}{Kamal, K.}, \bibinfo{author}{Farooq, B.}, \bibinfo{year}{2024}.
\newblock \bibinfo{title}{Ordinal-{ResLogit}: interpretable deep residual neural networks for ordered choices}.
\newblock \bibinfo{journal}{Journal of Choice Modelling} \bibinfo{volume}{50}, \bibinfo{pages}{100454}.
\newblock \DOIprefix\doi{10.1016/j.jocm.2023.100454}.
%Type = Inproceedings
\bibitem[{Kara{\c{c}}ay et~al.(2022)Kara{\c{c}}ay, Bianchi, G{\"u}nnemann and Bouchard}]{iohmm2022}
\bibinfo{author}{Kara{\c{c}}ay, B.}, \bibinfo{author}{Bianchi, M.}, \bibinfo{author}{G{\"u}nnemann, S.}, \bibinfo{author}{Bouchard, G.}, \bibinfo{year}{2022}.
\newblock \bibinfo{title}{Mixture of input-output hidden {Markov} models for heterogeneous disease progression modeling}, in: \bibinfo{booktitle}{Proceedings of the 1st Workshop on Healthcare AI and COVID-19, ICML 2022}.
%Type = Inproceedings
\bibitem[{Kendall and Gal(2017)}]{kendall2017uncertainties}
\bibinfo{author}{Kendall, A.}, \bibinfo{author}{Gal, Y.}, \bibinfo{year}{2017}.
\newblock \bibinfo{title}{What uncertainties do we need in {Bayesian} deep learning for computer vision?}, in: \bibinfo{booktitle}{Advances in Neural Information Processing Systems}, \bibinfo{publisher}{Curran Associates}.
%Type = Inproceedings
\bibitem[{Kingma and Ba(2015)}]{kingma2015adam}
\bibinfo{author}{Kingma, D.P.}, \bibinfo{author}{Ba, J.}, \bibinfo{year}{2015}.
\newblock \bibinfo{title}{Adam: a method for stochastic optimization}, in: \bibinfo{booktitle}{International Conference on Learning Representations}.
%Type = Inproceedings
\bibitem[{Kingma and Welling(2014)}]{kingma2014vae}
\bibinfo{author}{Kingma, D.P.}, \bibinfo{author}{Welling, M.}, \bibinfo{year}{2014}.
\newblock \bibinfo{title}{Auto-encoding variational {Bayes}}, in: \bibinfo{booktitle}{International Conference on Learning Representations}.
%Type = Article
\bibitem[{Koch et~al.(2024)Koch, Baumgartner and Berens}]{koch2024distribution}
\bibinfo{author}{Koch, L.M.}, \bibinfo{author}{Baumgartner, C.F.}, \bibinfo{author}{Berens, P.}, \bibinfo{year}{2024}.
\newblock \bibinfo{title}{Distribution shift detection for the postmarket surveillance of medical {AI} algorithms: a retrospective simulation study}.
\newblock \bibinfo{journal}{npj Digital Medicine} \bibinfo{volume}{7}, \bibinfo{pages}{113}.
\newblock \DOIprefix\doi{10.1038/s41746-024-01085-w}.
%Type = Article
\bibitem[{Kompa et~al.(2021)Kompa, Snoek and Beam}]{kompa2021second}
\bibinfo{author}{Kompa, B.}, \bibinfo{author}{Snoek, J.}, \bibinfo{author}{Beam, A.L.}, \bibinfo{year}{2021}.
\newblock \bibinfo{title}{Second opinion needed: communicating uncertainty in medical machine learning}.
\newblock \bibinfo{journal}{NPJ Digital Medicine} \bibinfo{volume}{4}, \bibinfo{pages}{4}.
\newblock \DOIprefix\doi{10.1038/s41746-020-00367-3}.
%Type = Inproceedings
\bibitem[{Lakshminarayanan et~al.(2017)Lakshminarayanan, Pritzel and Blundell}]{lakshminarayanan2017simple}
\bibinfo{author}{Lakshminarayanan, B.}, \bibinfo{author}{Pritzel, A.}, \bibinfo{author}{Blundell, C.}, \bibinfo{year}{2017}.
\newblock \bibinfo{title}{Simple and scalable predictive uncertainty estimation using deep ensembles}, in: \bibinfo{booktitle}{Advances in Neural Information Processing Systems}, \bibinfo{publisher}{Curran Associates}.
%Type = Article
\bibitem[{LaMontagne et~al.(2019)LaMontagne, Benzinger, Morris, Keefe, Hornbeck, Xiong, Grant, Hassenstab, Moulder, Vlassenko et~al.}]{lamontagne2019oasis}
\bibinfo{author}{LaMontagne, P.J.}, \bibinfo{author}{Benzinger, T.L.}, \bibinfo{author}{Morris, J.C.}, \bibinfo{author}{Keefe, S.}, \bibinfo{author}{Hornbeck, R.}, \bibinfo{author}{Xiong, C.}, \bibinfo{author}{Grant, E.}, \bibinfo{author}{Hassenstab, J.}, \bibinfo{author}{Moulder, K.}, \bibinfo{author}{Vlassenko, A.G.}, et~al., \bibinfo{year}{2019}.
\newblock \bibinfo{title}{{OASIS-3}: longitudinal neuroimaging, clinical, and cognitive dataset for normal aging and {Alzheimer's} disease}.
\newblock \bibinfo{journal}{medRxiv} \DOIprefix\doi{10.1101/2019.12.13.19014902}.
%Type = Article
\bibitem[{Lim et~al.(2021)Lim, Ar{\i}k, Loeff and Pfister}]{lim2021temporal}
\bibinfo{author}{Lim, B.}, \bibinfo{author}{Ar{\i}k, S.{\"O}.}, \bibinfo{author}{Loeff, N.}, \bibinfo{author}{Pfister, T.}, \bibinfo{year}{2021}.
\newblock \bibinfo{title}{Temporal fusion transformers for interpretable multi-horizon time series forecasting}.
\newblock \bibinfo{journal}{International Journal of Forecasting} \bibinfo{volume}{37}, \bibinfo{pages}{1748--1764}.
\newblock \DOIprefix\doi{10.1016/j.ijforecast.2021.03.012}.
%Type = Article
\bibitem[{Nguyen et~al.(2020)Nguyen, He, An, Alexander, Feng and Yeo}]{nguyen2020predicting}
\bibinfo{author}{Nguyen, M.}, \bibinfo{author}{He, T.}, \bibinfo{author}{An, L.}, \bibinfo{author}{Alexander, D.C.}, \bibinfo{author}{Feng, J.}, \bibinfo{author}{Yeo, B.T.T.}, \bibinfo{year}{2020}.
\newblock \bibinfo{title}{Predicting {Alzheimer's} disease progression using deep recurrent neural networks}.
\newblock \bibinfo{journal}{NeuroImage} \bibinfo{volume}{222}, \bibinfo{pages}{117203}.
\newblock \DOIprefix\doi{10.1016/j.neuroimage.2020.117203}.
%Type = Article
\bibitem[{Oxtoby et~al.(2018)Oxtoby, Young, Cash, Benzinger, Fagan, Morris, Bateman, Fox, Schott and Alexander}]{oxtoby2017data}
\bibinfo{author}{Oxtoby, N.P.}, \bibinfo{author}{Young, A.L.}, \bibinfo{author}{Cash, D.M.}, \bibinfo{author}{Benzinger, T.L.}, \bibinfo{author}{Fagan, A.M.}, \bibinfo{author}{Morris, J.C.}, \bibinfo{author}{Bateman, R.J.}, \bibinfo{author}{Fox, N.C.}, \bibinfo{author}{Schott, J.M.}, \bibinfo{author}{Alexander, D.C.}, \bibinfo{year}{2018}.
\newblock \bibinfo{title}{Data-driven models of dominantly-inherited {Alzheimer's} disease progression}.
\newblock \bibinfo{journal}{Brain} \bibinfo{volume}{141}, \bibinfo{pages}{1529--1544}.
\newblock \DOIprefix\doi{10.1093/brain/awy050}.
%Type = Article
\bibitem[{Petersen(2011)}]{petersen2014mild}
\bibinfo{author}{Petersen, R.C.}, \bibinfo{year}{2011}.
\newblock \bibinfo{title}{Mild cognitive impairment}.
\newblock \bibinfo{journal}{New England Journal of Medicine} \bibinfo{volume}{364}, \bibinfo{pages}{2227--2234}.
\newblock \DOIprefix\doi{10.1056/NEJMcp0910237}.
%Type = Inproceedings
\bibitem[{Phetrittikun and Suvirat(2023)}]{phetrittikun2023tft}
\bibinfo{author}{Phetrittikun, R.}, \bibinfo{author}{Suvirat, C.}, \bibinfo{year}{2023}.
\newblock \bibinfo{title}{Temporal fusion transformer for forecasting vital sign trajectories in intensive care patients}, in: \bibinfo{booktitle}{2023 IEEE International Conference on Electronics, Computing and Communication Technologies (CONECCT)}, \bibinfo{organization}{IEEE}. pp. \bibinfo{pages}{1--6}.
\newblock \DOIprefix\doi{10.1109/CONECCT57959.2023.10234585}.
%Type = Article
\bibitem[{Reiman et~al.(2011)Reiman, Langbaum, Fleisher, Caselli, Chen, Ayutyanont, Quiroz, Kosik, Lopera and Tariot}]{reiman2016alzheimer}
\bibinfo{author}{Reiman, E.M.}, \bibinfo{author}{Langbaum, J.B.}, \bibinfo{author}{Fleisher, A.S.}, \bibinfo{author}{Caselli, R.J.}, \bibinfo{author}{Chen, K.}, \bibinfo{author}{Ayutyanont, N.}, \bibinfo{author}{Quiroz, Y.T.}, \bibinfo{author}{Kosik, K.S.}, \bibinfo{author}{Lopera, F.}, \bibinfo{author}{Tariot, P.N.}, \bibinfo{year}{2011}.
\newblock \bibinfo{title}{Alzheimer's prevention initiative: a plan to accelerate the evaluation of presymptomatic treatments}.
\newblock \bibinfo{journal}{Journal of Alzheimer's Disease} \bibinfo{volume}{26}, \bibinfo{pages}{S321--S329}.
\newblock \DOIprefix\doi{10.3233/JAD-2011-0059}.
%Type = Book
\bibitem[{Rizopoulos(2012)}]{rizopoulos2012joint}
\bibinfo{author}{Rizopoulos, D.}, \bibinfo{year}{2012}.
\newblock \bibinfo{title}{Joint Models for Longitudinal and Time-to-Event Data: With Applications in {R}}.
\newblock \bibinfo{publisher}{CRC Press}, \bibinfo{address}{Boca Raton, FL}.
%Type = Article
\bibitem[{Shi et~al.(2023)Shi, Cao and Raschka}]{shi2023corn}
\bibinfo{author}{Shi, X.}, \bibinfo{author}{Cao, W.}, \bibinfo{author}{Raschka, S.}, \bibinfo{year}{2023}.
\newblock \bibinfo{title}{Deep neural networks for rank-consistent ordinal regression based on conditional probabilities}.
\newblock \bibinfo{journal}{Pattern Analysis and Applications} \bibinfo{volume}{26}, \bibinfo{pages}{941--955}.
\newblock \DOIprefix\doi{10.1007/s10044-023-01181-9}.
%Type = Article
\bibitem[{Tang et~al.(2025)Tang, Zhao, Chen, Liu and Zhang}]{tang2025mci}
\bibinfo{author}{Tang, X.}, \bibinfo{author}{Zhao, L.}, \bibinfo{author}{Chen, M.}, \bibinfo{author}{Liu, W.}, \bibinfo{author}{Zhang, J.}, \bibinfo{year}{2025}.
\newblock \bibinfo{title}{Predicting the progression of mild cognitive impairment based on fine-grained and spatiotemporal features of {MRI}}.
\newblock \bibinfo{journal}{Biomedical Signal Processing and Control} \bibinfo{volume}{98}, \bibinfo{pages}{107012}.
\newblock \DOIprefix\doi{10.1016/j.bspc.2025.107012}.
%Type = Inproceedings
\bibitem[{Vaswani et~al.(2017)Vaswani, Shazeer, Parmar, Uszkoreit, Jones, Gomez, Kaiser and Polosukhin}]{vaswani2017attention}
\bibinfo{author}{Vaswani, A.}, \bibinfo{author}{Shazeer, N.}, \bibinfo{author}{Parmar, N.}, \bibinfo{author}{Uszkoreit, J.}, \bibinfo{author}{Jones, L.}, \bibinfo{author}{Gomez, A.N.}, \bibinfo{author}{Kaiser, {\L}.}, \bibinfo{author}{Polosukhin, I.}, \bibinfo{year}{2017}.
\newblock \bibinfo{title}{Attention is all you need}, in: \bibinfo{booktitle}{Advances in Neural Information Processing Systems}, \bibinfo{publisher}{Curran Associates}.
%Type = Article
\bibitem[{Wang et~al.(2026)Wang, Liu and Fu}]{ordinaldr2026}
\bibinfo{author}{Wang, M.}, \bibinfo{author}{Liu, Y.}, \bibinfo{author}{Fu, H.}, \bibinfo{year}{2026}.
\newblock \bibinfo{title}{Uncertainty-aware ordinal deep learning for cross-dataset diabetic retinopathy grading}.
\newblock \bibinfo{journal}{arXiv preprint arXiv:2602.10315} .
%Type = Article
\bibitem[{Wang et~al.(2024)Wang, Gao, Wei, Johnston, Yuan, Zhang and Yu}]{wang2024multimodal}
\bibinfo{author}{Wang, Y.}, \bibinfo{author}{Gao, R.}, \bibinfo{author}{Wei, T.}, \bibinfo{author}{Johnston, L.}, \bibinfo{author}{Yuan, X.}, \bibinfo{author}{Zhang, Y.}, \bibinfo{author}{Yu, Z.}, \bibinfo{year}{2024}.
\newblock \bibinfo{title}{Predicting long-term progression of {Alzheimer's} disease using a multimodal deep learning model incorporating interaction effects}.
\newblock \bibinfo{journal}{Journal of Translational Medicine} \bibinfo{volume}{22}, \bibinfo{pages}{245}.
\newblock \DOIprefix\doi{10.1186/s12967-024-05025-w}.
%Type = Article
\bibitem[{Weiner et~al.(2017)Weiner, Veitch, Aisen, Beckett, Cairns, Cedarbaum, Donohue, Green, Harvey, Jack et~al.}]{weiner2017alzheimer}
\bibinfo{author}{Weiner, M.W.}, \bibinfo{author}{Veitch, D.P.}, \bibinfo{author}{Aisen, P.S.}, \bibinfo{author}{Beckett, L.A.}, \bibinfo{author}{Cairns, N.J.}, \bibinfo{author}{Cedarbaum, J.}, \bibinfo{author}{Donohue, M.C.}, \bibinfo{author}{Green, R.C.}, \bibinfo{author}{Harvey, D.}, \bibinfo{author}{Jack, C.R.}, et~al., \bibinfo{year}{2017}.
\newblock \bibinfo{title}{The {Alzheimer's Disease Neuroimaging Initiative 3}: continued innovation for clinical trial improvement}.
\newblock \bibinfo{journal}{Alzheimer's \& Dementia} \bibinfo{volume}{13}, \bibinfo{pages}{561--571}.
\newblock \DOIprefix\doi{10.1016/j.jalz.2016.10.006}.
%Type = Article
\bibitem[{Weng et~al.(2025)Weng, Liu, Huang, Hsieh and Foschini}]{weng2025ood}
\bibinfo{author}{Weng, W.H.}, \bibinfo{author}{Liu, Q.}, \bibinfo{author}{Huang, R.}, \bibinfo{author}{Hsieh, J.}, \bibinfo{author}{Foschini, L.}, \bibinfo{year}{2025}.
\newblock \bibinfo{title}{First, do no harm: addressing {AI}'s challenges with out-of-distribution data in medicine}.
\newblock \bibinfo{journal}{Clinical and Translational Science} \bibinfo{volume}{18}, \bibinfo{pages}{e70132}.
\newblock \DOIprefix\doi{10.1111/cts.70132}.
%Type = Techreport
\bibitem[{{World Health Organization}(2023)}]{who2023dementia}
\bibinfo{author}{{World Health Organization}}, \bibinfo{year}{2023}.
\newblock \bibinfo{title}{Dementia}.
\newblock \bibinfo{type}{Technical Report}. World Health Organization.
\newblock \bibinfo{note}{Fact sheet. Available at: \url{https://www.who.int/news-room/fact-sheets/detail/dementia}}.
%Type = Article
\bibitem[{Zhang et~al.(2025)Zhang, Chen, Hern{\'a}ndez-Lobato and Li}]{uqhealthcare2025}
\bibinfo{author}{Zhang, Z.}, \bibinfo{author}{Chen, T.}, \bibinfo{author}{Hern{\'a}ndez-Lobato, J.M.}, \bibinfo{author}{Li, S.}, \bibinfo{year}{2025}.
\newblock \bibinfo{title}{Uncertainty quantification for machine learning in healthcare: a survey}.
\newblock \bibinfo{journal}{arXiv preprint arXiv:2505.02874} .
%Type = Article
\bibitem[{Zhao et~al.(2024)Zhao, Guo, Wang and Shen}]{oodmedical2024}
\bibinfo{author}{Zhao, T.}, \bibinfo{author}{Guo, Y.}, \bibinfo{author}{Wang, X.}, \bibinfo{author}{Shen, D.}, \bibinfo{year}{2024}.
\newblock \bibinfo{title}{Out-of-distribution detection in medical image analysis: a survey}.
\newblock \bibinfo{journal}{arXiv preprint arXiv:2404.18279} .

\end{thebibliography}

\end{document}